\documentclass[journal=jacsat,manuscript=article]{achemso}
\usepackage{comment}
\usepackage{graphicx}
\SectionNumbersOn
\usepackage{graphicx}
\usepackage{tabularx}
\usepackage{gensymb}
\usepackage{subfig}
\usepackage{csvsimple}
\usepackage[figuresright]{rotating}
\usepackage{multirow}
\usepackage[export]{adjustbox}
\usepackage{subfig}
\usepackage{lineno}
\usepackage[export]{adjustbox}
\usepackage{hyperref}
\hypersetup{
    colorlinks=true,
    linkcolor=blue,
    citecolor=blue,
    urlcolor=blue
}
\usepackage{chemformula} 
\usepackage[T1]{fontenc} 
\author{Suchetana Sadhukhan}
\affiliation{School of Advanced Sciences and Languages, VIT Bhopal University, Kothri Kalan, Sehore-466114, Madhya Pradesh, India}

\author{Vivek Kumar Yadav}
\affiliation{Department of Chemistry, University of Allahabad, U.P.- 211001, Uttar Pradesh, India}
\email{vkyadav@allduniv.ac.in}

\title{Forecasting, capturing and activation of carbon-dioxide ($CO_{2}$): Integration of Time Series Analysis, Machine Learning, and Material Design}

\begin{document}

\begin{tocentry}
\includegraphics[width=6cm]{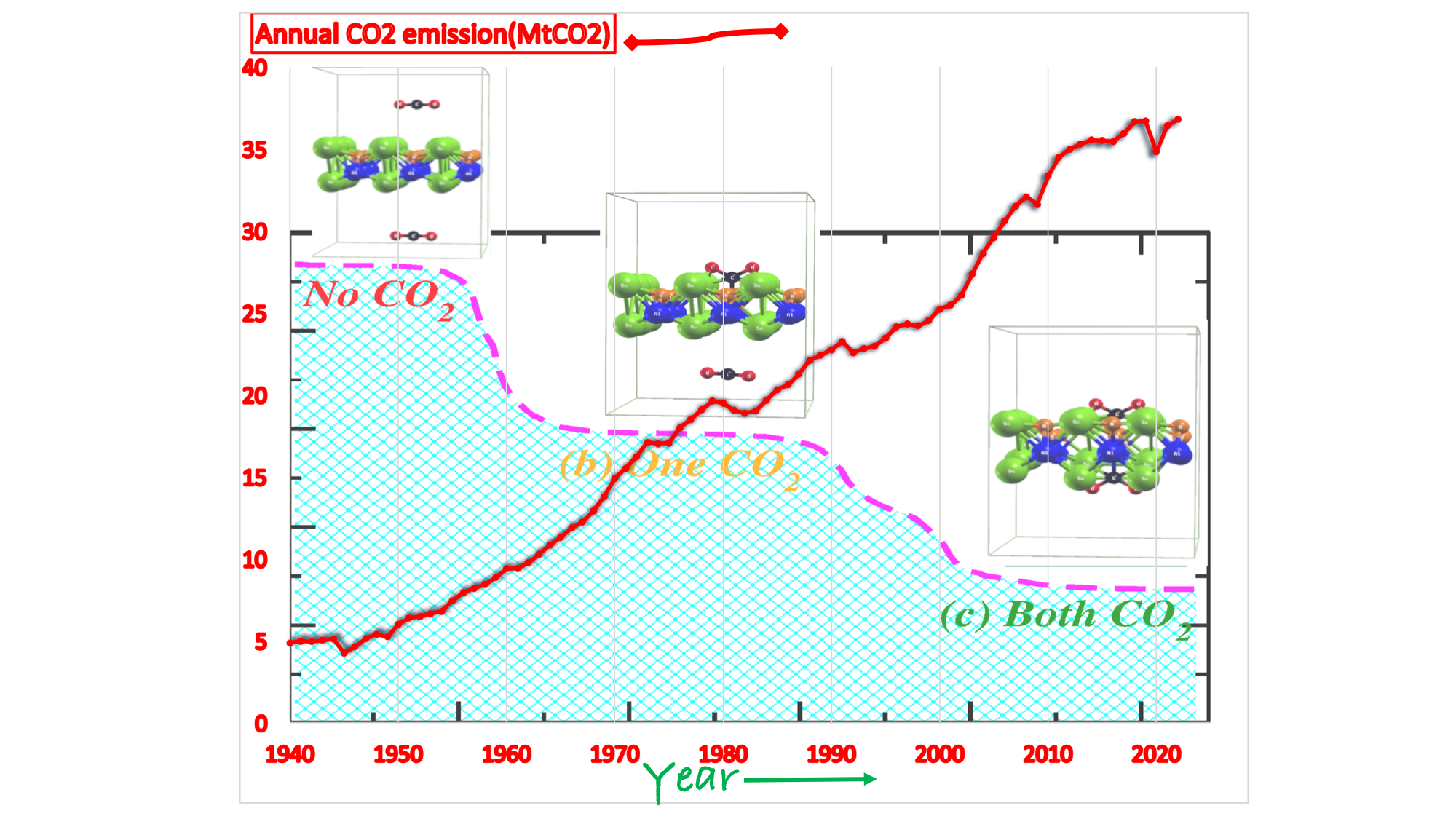}
\end{tocentry}

\begin{abstract}
This study provides a comprehensive time series analysis of daily industry-specific, country-wise $CO_2$ emissions from January $2019$ to February $2023$. The research focuses on the Power, Industry, Ground Transport, Domestic Aviation, and International Aviation sectors in European countries (EU27 \& UK, Italy, Germany, Spain) and India, utilizing near-real-time activity data from the Carbon Monitor research initiative. To identify regular emission patterns, the data from the year 2020 is excluded due to the disruptive effects caused by the COVID-19 pandemic. The study then performs a principal component analysis (PCA) to determine the key contributors to $CO_2$ emissions. The analysis reveals that the Power, Industry, and Ground Transport sectors account for a significant portion of the variance in the dataset. A 7-day moving averaged dataset is employed for further analysis to facilitate robust predictions. This dataset captures both short-term and long-term trends and enhances the quality of the data for prediction purposes. The study utilizes Long Short-Term Memory (LSTM) models on the $7$-day moving averaged dataset to effectively predict emissions and provide insights for policy decisions, mitigation strategies, and climate change efforts. During the training phase, the stability and convergence of the LSTM models are ensured, which guarantees their reliability in the testing phase. The evaluation of the loss function indicates this reliability. The model achieves high efficiency, as demonstrated by $R^2$ values ranging from $0.8242$ to $0.995$ for various countries and sectors. Furthermore, there is a proposal for utilizing scandium and boron/aluminium-based thin films as exceptionally efficient materials for capturing $CO_2$ (with a binding energy range from -$3.0$ to -$3.5$ eV). These materials are shown to surpass the affinity of graphene and boron nitride sheets in this regard.
\end{abstract}

\section*{Keywords}
$CO_2$ emissions and capture, Time series analysis, Principal Component Analysis (PCA), Long Short-Term Memory (LSTM) models, DFT calculations, Material design

\section*{Synopsis}
This study provides a comprehensive time series analysis of daily industry-specific, country-wise $CO_2$ emissions, identifying key contributors and proposing efficient $CO_2$ capture materials.

\section{Introduction}
The global population is estimated to have surpassed 8 billion people by November $15$, $2022$, indicating rapid growth that has posed significant challenges in terms of global Carbon-di-Oxide ($CO_2$) emissions \cite{UNPopulation2022}. This exponential population growth has contributed to a surge in $CO_2$ emissions, presenting severe environmental challenges worldwide \cite{bonga2014level}. Global $CO_2$ emissions have been rising at a rate higher than the global average, leading to widespread consequences such as food and water crises, as well as heightened frequency and intensity of natural disasters \cite{olivier2017trends}. The health impacts of $CO_2$ emissions are diverse, encompassing direct effects like respiratory illnesses and vision impairment, as well as indirect effects such as climate change and global warming \cite{augbulut2022forecasting, bakay2021electricity}. Furthermore, air pollution stemming from $CO_2$ emissions results in severe diseases and a significant number of annual deaths across the globe \cite{augbulut2022forecasting}. The escalating number of vehicles on the roads worldwide is a major contributor to the rising levels of $CO_2$ emissions \cite{augbulut2022forecasting, ahmadi2019environmental}. An average temperature increase of $0.6$ degrees Celsius has been observed in the Earth's climate during the $20$-th century, with an indication of further rise in the coming century \cite{bistline2010role}.  
If anthropogenic $CO_2$ emissions continue unabated, the average global temperature could easily surpass a $2$-degree Celsius increase in the near future.
It is imperative to address the relentless growth of global $CO_2$ emissions, as they serve as a major contributor to global warming, leading to substantial environmental, social, and economic threats worldwide. Policymakers must take decisive actions to tackle these critical issues. In this line, accurate forecasting of $CO_2$ emissions plays a crucial role in creating public awareness and developing effective strategies to mitigate the adverse effects. Various statistical, machine learning and deep learning models have been employed to forecast $CO_2$ emissions using historical data spanning several decades. This paper aims to gain insights into the current trajectory of global $CO_2$ emissions and identify necessary corrective measures to control $CO_2$ emissions.

This paper addresses three main areas: 1) to identify influential factors, 2) to forecast emissions, and 3) to explore Carbon Capture, Utilization, and Storage (CCUS). Various well-established methods are available to extract information about influential factors. The STIRPAT model has been employed in the study by Wen et al. to investigate the key drivers of urban residential energy consumption and $CO_2$ emissions in China, revealing that factors such as population magnitude, wealth, and population density contribute to increased energy consumption and $CO_2$ emissions \cite{wen2020influencing}. The index decomposition analysis (IDA) approach has been employed to understand the impact of international trade on global $CO_2$ emissions \cite{wang2018assessing}. It allows for the quantification of trade-related factors such as embodied emissions in trade, carbon leakage, and emission transfers. Donglan et al. (2010) have utilized IDA to investigate the underlying drivers of residential carbon dioxide ($CO_2$) emissions in urban and rural China \cite{donglan2010driving}. By using the IDA approach, they have been able to quantify diverse factors, such as population, urbanization, income, energy intensity, household size, and lifestyle, thereby discerning their respective contributions to emissions. Liu et al. (2019) have focused on the Extended Logarithmic Mean Division Index Decomposition (ELMDI) method to analyze carbon dioxide ($CO_2$) emissions in China's manufacturing industry \cite{liu2019analysis}. The ELMDI approach comprehensively assesses the driving factors behind $CO_2$ emissions, including energy intensity, emission coefficient, structural effect, and scale effect. Gonzalez et al. (2014) have presented the Logarithmic-Mean Divisia Index (LMDI) decomposition method with the activity revaluation approach to track carbon dioxide ($CO_2$) emissions in the European Union (EU) \cite{gonzalez2014tracking}. Wang et al. (2005) examined energy-related carbon dioxide ($CO_2$) emissions in China from 1957 to 2000 using decomposition analysis \cite{wang2005decomposition}. Input-output structural decomposition analysis (SDA) is an economic analysis method used to assess the drivers of environmental impacts, particularly concerning carbon dioxide ($CO_2$) emissions. SDA utilizes input-output tables, which provide a comprehensive account of economic activities and transactions between sectors \cite{wang2013carbon}. In a study by Wang et al. (2013), energy-related $CO_2$ emissions in Beijing from 2000 to 2010 were examined using input-output and structural decomposition analyses. The findings reveal that sectoral $CO_2$ emissions increased due to economic growth and structural changes, emphasizing the need to optimize the economic structure, improve technology, and explore new energy sources to reduce emissions and achieve sustainable development \cite{wei2017driving}. The refined Laspeyres method has been employed to analyze the factors driving $CO_2$ emission growth, offering valuable insights into sectoral policies and their impacts on emissions \cite{kumbarouglu2011sectoral}. Ang (1997) proposed a refined Divisia index method to overcome problems in decomposing energy consumption or emissions in the industry, ensuring a perfect decomposition without residuals and addressing issues with zero values in the dataset \cite{ang1997decomposition}. Ang (2003) compared the decomposition technique proposed by Albrecht et al. (2002) with the method by Sun (1998), finding them to be exactly the same, and provides a comprehensive overview of optimal decomposition methodologies and their importance in energy demand and related analysis \cite{ang2003perfect}. Comprehensive country studies have explored the relationships among $CO_2$ emissions, energy utilization, economic expansion, and trade accessibility, indicating support for the Environmental Kuznets Curve hypothesis and highlighting the role of economic factors in emissions \cite{shahbaz2012environmental}. Jalil (2009) explored the long-term correlation between carbon emissions, energy usage, revenue, and international trade in China, finding a quadratic relationship between income and $CO_2$ emissions supporting the Environmental Kuznets Curve (EKC) \cite{jalil2009environment}. Studies have analyzed the long-run dynamics and causal relationships between carbon emissions, energy consumption, and industrial growth using co-integration analysis, providing evidence of significant impacts on carbon emissions in both short and long runs \cite{rahman2017carbon}. Each of these methods offers valuable insights into understanding and quantifying the influential factors contributing to emissions. In this paper, principal component analysis (PCA) is utilized to merge the original features, reduce dimensionality, and identify the major contributors \cite{pearson1901liii, hotelling1933analysis}. By doing so, it simplifies computation and enables the selection of appropriate inputs for the forecasting model.

The literature on forecasting $CO_2$ emissions encompasses various models, starting with traditional approaches such as multiple linear regression and multiple polynomial regression \cite{ciulla2019building, hosseini2019forecasting, ostertagova2012modelling}. These methods have established relationships between $CO_2$ emissions and independent variables. However, they often encounter limitations due to the complex and dynamic nature of $CO_2$ emissions. To overcome the limitations of traditional prediction methods, researchers have turned to artificial intelligence technology and machine learning algorithms. In recent years, swarm optimization algorithms and neural network models have gained significant attention \cite{qiao2020hybrid, sun2017factor, gallo2014neural, pino2017comparison}. The Particle Swarm Optimization-Back Propagation Neural Network (PSO-BPNN) model has been proposed to optimize the weights and biases of the neural network, enhancing the accuracy of $CO_2$ emission predictions\ cite{sun2016using}. Another approach, the optimized Extreme Learning Machine (ELM), has been used to forecast $CO_2$ emissions by optimizing the hidden layer parameters of the ELM algorithm \cite{sun2017factor}. Among the machine learning models, Long Short-Term Memory (LSTM) has emerged as a powerful technique for $CO_2$ emission forecasting \cite{liu2018hybrid, zuo2020lstm, zhu2022lstm, kumari2022machine, liu2020carbon, ke2023carbon}. LSTM models are capable of capturing long-term dependencies and temporal patterns in data, making them particularly suitable for time series forecasting \cite{kumari2022machine, faruque2022comparative, zheng2019spatial, zuo2020lstm}. The advantages of LSTM models include their ability to handle non-linear relationships, capture complex dynamics, and effectively model both short and long-term dependencies.
Recently, there have been numerous proposals for $2$D transition metal-containing
sheets, which exhibit enhanced catalytic activity for $CO_2$ adsorption and other
flue gases found in the environment. Two valuable materials, namely MXene and
MBenes have been synthesized theoretically and experimentally from the MAB
phase (where M is a transition metal, and A is a group IIIA or IVA element) \cite{VKY0, VKY1, VKY2, VKY3, VKY4}.
A new family of 2D transition metal borides/aluminium, similar to the well-
established MXene family can be generated by selectively etching out either
element A or B (such as boron or aluminium). Wang and coworkers \cite{VKY5}
demonstrate using $Ti_3C_2T_x $-based MXene nanosheets as catalysts in
photocatalytic $CO_2$ reduction, with significantly higher yield than bulk $Ti_3C_2T_x$
MXene powder. This work also opens up possibilities for mass-producing MXene
nanosheets with highly active photocatalysis.
Other MBenes, such as $Cr_2B_2$ and $Mo_2B_2$, have also been synthesized
experimentally from their respective MAB phases. Unlike MXenes, all MBenes can
be stabilized without surface passivation groups, making them ideal platforms for
exploring the catalytic behavior of boron-containing materials.
Mou and coworkers \cite{VKY6} recently successfully synthesized the $Mo_2AlB_2$
compound and 2D MoB nanosheets through a $ZnCl_2$ molten salt etching approach
at relatively low temperatures. Their work confirms that the MoB MBene can be
prepared by etching the as-synthesized $Mo_2AlB_2$ precursor in a LiF–HCl solution.
Weerasinghe et al. \cite{VKY7} exfoliated 2D mica nanosheets (eMica nanosheets) and
characterized them using various techniques. They demonstrated the ability of
eMica nanosheets to capture $CO_2$, showing an 87 percent increase in $CO_2$
adsorption capacity compared to conventional mica.
Computational studies by Xiao and Shen \cite{VKY8} investigated the catalytic
performance of a series of M2B2-type MBenes to reduce nitric oxide (NO) to $NH_3$.
Their work showed excellent catalytic performance and lower limiting potential
for NO to $NH_3$ conversion.
For $CO_2$ capture and activation, we selected two different MBenes based on
Scandium (Metal) and boron/aluminium. The ideal substrate would activate $CO_2$
through charge transfer, resulting in a bent anionic $CO_2$ moiety, which is more
reactive when combined with other surface chemicals.

The paper is structured into three main sections. Section \ref{met} consists of a dataset description, proposed methodologies, performance metrics analysis, and hyperparameter setting, including a detailed discussion on the suitable material design aspects. Section \ref{res} focuses on the results and analysis derived from the study. Finally, in Section \ref{con}, the paper concludes by summarizing the findings and implications of the research.

\section{Materials and Methods}\label{met}
\subsection{Data Analysis}
In this study, we conduct a comprehensive time series analysis of daily industry-specific, country-wise $CO_2$ emissions from January $1^{st}$, $2019$ to February $28^{th}$, $2023$. Our analysis focuses on the Power, Industry, Ground Transport, Domestic Aviation, and International Aviation sectors, examining the emissions data for a combination of European countries and major economies, namely EU27 \& UK, Italy, Germany, and Spain, as well as India. The data utilized in this study is primarily sourced from near-real-time activity data, which is publicly available through the \href{https://carbonmonitor.org/}{Carbon Monitor research initiative}.  
\subsection{Principal Component Analysis}
Principal Component Analysis(PCA) is a powerful technique commonly employed in time series analysis and machine learning for dimensionality reduction \cite{pearson1901liii, hotelling1933analysis}. It allows for the transformation of high-dimensional data into a lower-dimensional space by identifying the directions of maximum variance and projecting the data in these directions. PCA determines a set of orthogonal vectors, known as principal components, that capture the highest variance in the data. These principal components are derived from the eigenvectors $O$ of the covariance matrix $C= \frac{1}{n} \sum_{i=1}^{n} x_i {x_i}^T$ where  $\mathbf{x}_i = (x_i(1), x_i(2),..., x_i(n))^T$ represents the data of length $n$. The covariance matrix describes the relationships between different dimensions.
\begin{equation}
\mathbf{y}_i=O^T x_i
\end{equation}
The eigenvectors with the largest eigenvalues correspond to the directions of maximum variance in the data, making them essential for capturing the underlying data structure. By projecting the original data onto the principal components, PCA effectively transforms the data into a new coordinate system where the dimensions are uncorrelated and ordered by their importance. PCA finds broad applications in various fields, including image and signal processing, bioinformatics, and finance. It is particularly useful for data visualization, as it enables the reduction of high-dimensional data to two or three dimensions, facilitating easy visualization and interpretation.
\subsection{Long Short-Term Memory}
The Long Short-Term Memory (LSTM) neural network is a specialized type of recurrent neural network (RNN) that effectively addresses the vanishing gradient problem encountered in traditional RNNs. This issue arises when gradients diminish exponentially as they propagate backwards in time, hindering the network's ability to learn long-term dependencies in sequential data. LSTM networks were initially introduced in 1997 by Hochreiter and Schmidhuber \cite{hochreiter1997long} and have subsequently gained popularity for modelling sequential data \cite{van2020review, graves2012long}.

\begin{figure}[hbtp]
\begin{center}
\includegraphics[width=8cm]{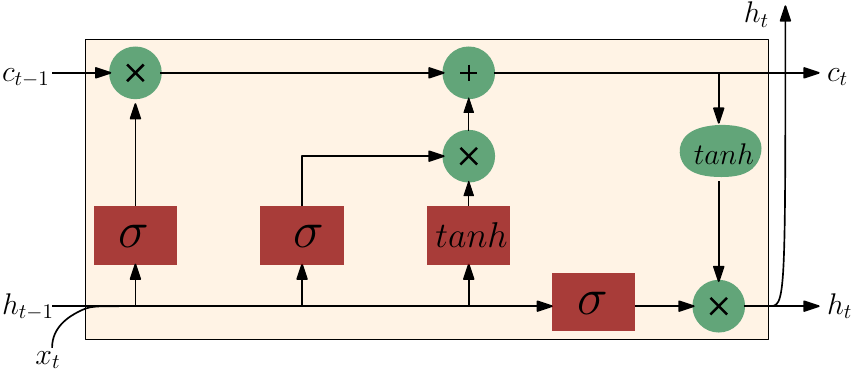}
\end{center}
\caption{LSTM structure}
\label{fig1}
\end{figure}

The architecture of an LSTM network consists of a sequence of LSTM cells, each featuring a memory cell responsible for storing and retrieving information across time steps. The detailed architecture is provided in Fig. \ref{fig1}. Additionally, there are three essential gating units within each LSTM cell: the input gate, forget gate, and output gate. The input gate regulates the flow of relevant information into the memory cell, while the forget gate determines which information is no longer useful and should be discarded. The output gate regulates the transmission of information from the memory cell to the output, ultimately used for making predictions. During training, the LSTM neural network learns to optimize a given objective function by adjusting the parameters associated with the gating units and memory cells.

\subsubsection{Performance Evaluating Index}
When assessing the performance of LSTM models, it is crucial to utilize appropriate evaluation metrics. This response provides a comprehensive discussion of commonly used evaluation metrics for evaluating LSTM model performance.
\begin{itemize}
\item Mean Squared Error (MSE): MSE is a widely employed metric in regression tasks, including LSTM models. It quantifies the mean square deviation between the predicted and observed values \cite{augbulut2021prediction, kumar2020time}. The MSE formula is expressed as:

\begin{equation}
MSE = \frac{1}{n} \sum_{i=1}^{n} (y_{\text{A}} - y_{\text{P}})^2
\end{equation}

Here, $y_{\text{A}}$ represents the actual value, $y_{\text{P}}$ denotes the predicted value, and $n$ signifies the number of datapoints. Lower values of MSE indicate superior performance \cite{bakay2021electricity, augbulut2021experimental}.
\item Root Mean Squared Error (RMSE): RMSE is another widely adopted regression metric that shares similarities with MSE. However, it takes the square root of MSE to yield a value in the same units as the target variable \cite{kumar2020time}. The RMSE formula is:

\begin{equation}
RMSE = \sqrt{MSE}
\end{equation}

Lower RMSE values correspond to better performance \cite{bakay2021electricity, augbulut2021experimental}..
\item Mean Absolute Error (MAE): MAE assesses the mean absolute difference between the predicted and original values. It is more robust to outliers compared to MSE and RMSE. The MAE formula is:

\begin{equation}
\text{{MAE}} = \frac{1}{n} \sum_{i=1}^{n} \lvert y_{\text{{A}}} - y_{\text{{P}}} \rvert
\end{equation}

Lower MAE values indicate improved performance.
\item R-Squared $(R^2)$: $R^2$ is a metric that quantifies the proportion of variance in the target variable explained by the LSTM model. It ranges between 0 and 1, with higher values indicating superior performance \cite{kumar2020time}. The $R^2$ formula is:

\begin{equation}
R^2 = 1 - \frac{SS_{\text{R}}}{SS_{\text{T}}}
\end{equation}

Here, $SS_{\text{R}}$ denotes the sum of squared residuals $=\sum_{i=1}^{n} (y_{\text{A}} - y_{\text{P}})^2$, and $SS_{\text{T}}$ represents the total sum of squares $=\sum_{i=1}^{n} (y_{\text{A}} - \langle y_{\text{A}} \rangle)^2$. Here $\langle ... \rangle$ denotes the mean value. Higher $R^2$ values correspond to better performance \cite{augbulut2021experimental}.
\end{itemize}
These evaluation metrics provide valuable insights into the performance of LSTM models and aid in determining their effectiveness in prediction.
\subsubsection{Hyperparameter Setting}
Optimal hyperparameter settings play a crucial role in constructing an LSTM model, as they significantly impact the model's performance. Key hyperparameters such as the number of LSTM layers and units, dropout rate, learning rate, batch size, sequence length, activation functions, and optimization algorithm require careful tuning. Adjusting these hyperparameters appropriately is essential for achieving optimal model performance \cite{faruque2022comparative}. Increasing the number of LSTM layers and units can enhance the model's performance by enabling it to learn more complex patterns. However, an excessive number of layers and units may lead to over-fitting, where the model becomes overly customized to the training data and performs inadequately on unseen data. Additionally, a large number of layers and units can result in slower training times, necessitating a trade-off between performance and computational efficiency. The dropout rate hyperparameter serves as a regularization technique to prevent over-fitting. It controls the percentage of LSTM units that are randomly dropped during training, forcing the model to rely on different combinations of units for better generalization. Tuning the dropout rate involves striking a balance between preventing over-fitting and avoiding under-fitting, where too high a dropout rate can hinder the model's learning capabilities. The learning rate hyperparameter determines the step size taken by the optimizer during training. Setting an appropriate learning rate is crucial for achieving convergence within a reasonable number of training iterations. A learning rate that is too high may result in overshooting and instability, while a learning rate that is too low can cause slow convergence. Finding the optimal learning rate requires careful experimentation and monitoring of the training process. Other hyperparameters such as sequence length, activation functions, and optimization algorithm also impact the model's performance. The sequence length determines the number of time steps considered as input for each training instance. Choosing an appropriate sequence length is important to capture the relevant temporal dependencies in the data. The activation functions, such as the sigmoid or tanh functions, introduce non-linearity to the LSTM units and influence the model's expressive power. The optimization algorithm, such as Adam or RMSprop, determines the approach employed to update the model's parameters during the training process.

In the specific case of this study, the units hyperparameter is set to $50$, specifying the number of LSTM units in each layer. The return sequences parameter is set to True for the first two LSTM layers, ensuring that the output of all LSTM layers is passed to the subsequent layers. The last LSTM layer, however, does not need to return sequences, so it is set to False. The model employs the Adam optimizer and utilizes the Mean Squared Error (MSE) loss function. The remaining hyperparameters are tuned individually for each dataset, with detailed discussions provided in Section \ref{res} and Table \ref{tab2}. 
\subsection{Computational Design of Two-Dimensional Material}
\begin{figure}[htbp]
\begin{center}
\includegraphics[width=8cm]{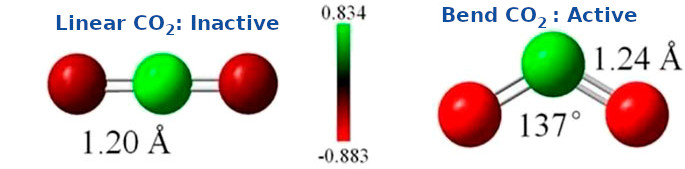}
\end{center}
\caption{ The molecular structure of linear and bend $CO_2$. The linear $CO_2$ is less strained compared to the bend one. }
\label{fig2}
\end{figure}

To overcome the issue of growing $CO_2$ concentration globally, as depicted by time series analysis and machine learning forecasting, efforts have been made to evict the gas from the environment and convert it into another useful product. The best way is to design new materials that efficiently capture the flue gas, especially $CO_2$, and convert it into useful products like methanol, methane etc, via $CO_2$ activation. The $CO_2$ activation requires conversion of its stable linear form to fragile bend form as shown in Fig. \ref{fig2}\\

\begin{figure}[htbp]
\begin{center}
\includegraphics[width=8cm,height=6cm]{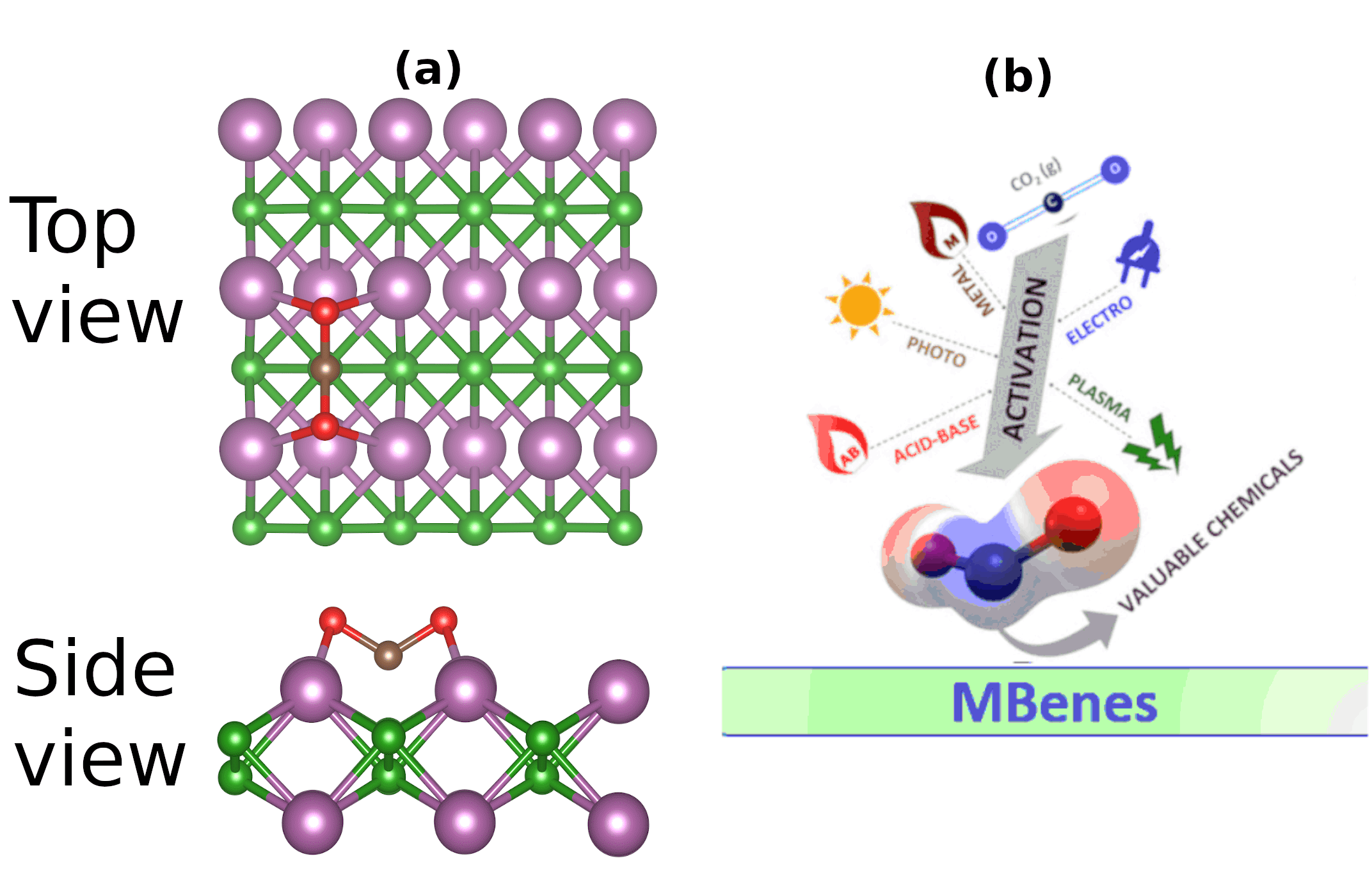}
\end{center}
\caption{(a) Two-dimensional MBene sheet for adsorption and activation of $CO_2$. (b) Once the $CO_2$ is activated, it can be further used in different useful products via catalysis.}
\label{fig3}
\end{figure}

Density functional theory {DFT} is a tool to determine the ground state and excited-state properties of a system like an atom, molecule, cluster, or solid. The basic principle of the theory is that any property of a many-electron system can be regarded as a function of electron density. DFT is a complementary and alternate approach to the traditional wave function-based methods of quantum physics and chemistry, which is represented by the many-electron wave function. In this work, we will use DFT as implemented in Quantum Espresso \cite{VKY9} simulation package to calculate the structural and electronic properties of the newly designed 2D thin films that will efficiently capture and activate $CO_2$. A benchmark calculation will be performed to obtain the reasonable plane wave energy cut-off and K-points for Brillouin zone sampling. To optimize the crystal geometry and predict the ground state properties, we will employ the generalized gradient approximation \cite{VKY10} (GGA) functional with PBE exchange-correlation \cite{VKY10}, which yields reasonably accurate ground state properties. Using GGA functional, the cohesive energy of these materials will be calculated to examine their thermodynamic stability. The Grimme’s D3 approach to include such interactions in our calculations \cite{VKY11}. The geometry optimization utilized a k-mesh of 10 × 10 × 1, while the self-consistent field (SCF) calculations were conducted on a K-point grid of 20 × 20 × 1, employing the Monkhorst–Pack \cite{VKY12} scheme. The cohesive energy of 2D material will be calculated using the equation below.
\begin{equation}
        E_{coh} = (E_{system}  - n \sum E_{x})/N
        \label{eqn6}
\end{equation}
where $E_{system}$ is the energy of two-dimensional material, $E_{x}$ is the energy of constituents, and n represents the total number of atoms in the system and N in bulk.
The substrate and adsorbate's binding energy ($E_{b}$) will be calculated using the equation below.
\begin{equation}
        E_{b} = E_{total} - (E_{substrate}+E_{adsorbate})
        \label{eqn7}
\end{equation}
where $E_{total}$ is the system's total energy, $E_{substrate}$ and $E_{adsorbate}$ are the substrate and adsorbate energy, respectively. Fig. \ref{fig3}(a) below shows the expected adsorption of $CO_2$ over computationally designed 2D materials designed by using electron-rich metal (Scandium) and electron-deficient non-metal (Boron) for efficient capture and activation. This combination yields a new class of two-dimensional materials, namely MBenes, which shows excellent adsorption and catalytic activity towards industrial flue gases (namely $CO_2$, $SO_2$, $O_3$, $NO_2$) as shown in Fig. \ref{fig3}(b).

In our current research, we are conducting ab initio Density Functional Theory (DFT) calculations to investigate the adsorption of individual molecules on the surfaces of three newly designed two-dimensional sheets. The unit cell of these sheets, namely $Sc_{18}Al_{1}B_{17}$ (SAB) and $Sc_{18}Al_{18}$ (SA), consists of a total of 36 atoms. Due to the significant computational complexity involved, we are constrained to conducting smaller-scale calculations. In the SAB and SA sheets, we are adsorbing/capturing only one $CO_2$ molecule from one side. However, in the $Sc_{18}Al_{9}B_{9}$ sheet, we have demonstrated dual-side adsorption, making this particular proposed sheet more efficient.

\section{Results and Discussion} \label{res}
In this section, we delve into a detailed integrated study that combines time series analysis, machine learning techniques, and computational design of two-dimensional materials specifically tailored for capturing and removing $CO_2$ gas. By synergizing time series analysis and advanced machine learning models, we gain a comprehensive understanding of $CO_2$ emission dynamics over time and accurate predictions for the future. Additionally, we investigate novel two-dimensional materials' potential to efficiently capture and remove $CO_2$ from the atmosphere, a promising avenue for sustainable climate change mitigation.
\subsection{Time Series Analysis}
\begin{figure}[htbp]
\begin{center}
\includegraphics[width=8cm]{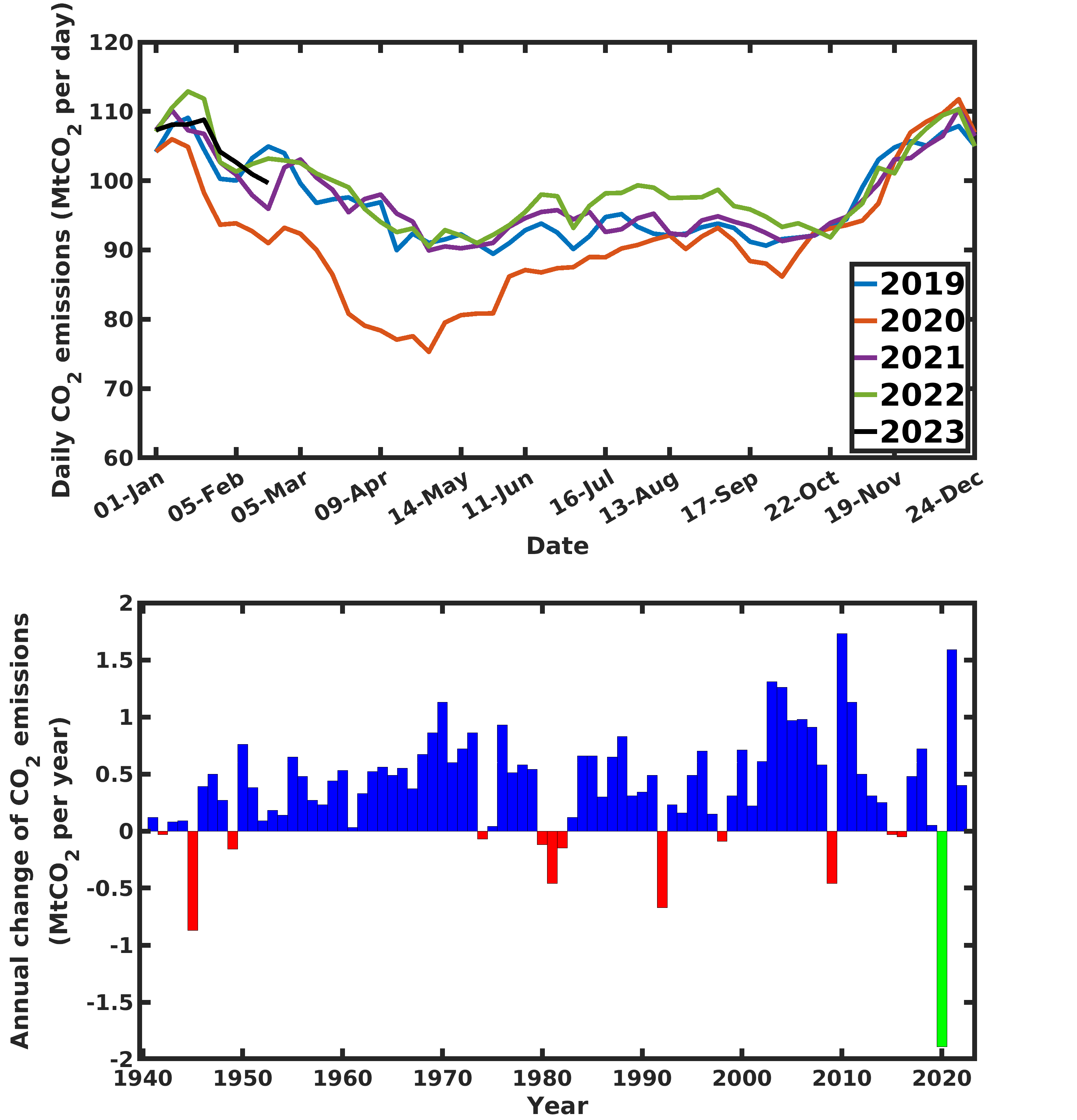}
\end{center}
\caption{(Top) Weekly averaged total emissions of $CO_2$ globally from $2019$ to $2023$ (as of $28$-th February $2023$) (Bottom) Annual changes in global $CO_2$ emissions from $1940$ to $2022$. Each bar represents a specific year, and the height of the bar indicates the magnitude of the annual change in emissions. The green bar specifically represents the change in emissions from $2020$ to $2021$, highlighting the shift in emissions levels following the COVID-$19$ pandemic.}
\label{fig4}
\end{figure}

To analyze the overall effect of country-wise total emissions, we aggregate the daily data from all sectors to obtain country-wise data. 
These estimates provide insights into the dynamics of $CO_2$ emissions before and after the COVID-$19$ pandemic, as well as the environmental impact triggered by the pandemic. 
Figure \ref{fig4} (top) illustrates the $7$-day averaged global $CO_2$ emissions computed from the normalized dataset. The graph clearly depicts the disparity in emissions between the pandemic year $(2020)$ and the other years $(2019$, $2021$, $2022$ and $2023)$. Following the pandemic year, emissions gradually returned to pre-pandemic levels. The bar plot in Fig. \ref{fig4} (bottom) provides a comprehensive overview of the annual changes in global $CO_2$ emissions from $1940$ to $2022$. Despite experiencing a significant decline during the pandemic, emissions swiftly rebounded to their pre-pandemic levels in $2021$. 

It is important to note that the COVID-$19$ pandemic caused widespread disruptions in economic activities, travel restrictions, and changes in energy consumption patterns. These unique circumstances led to anomalous trends in $CO_2$ emissions that are unlikely to be representative of typical long-term patterns. Therefore, for further analysis, we exclude the $2020$ data in order to focus on capturing and predicting the more regular and predictable emission patterns observed in other years.

\begin{figure}[htbp]
\begin{center}
\includegraphics[width=10cm]{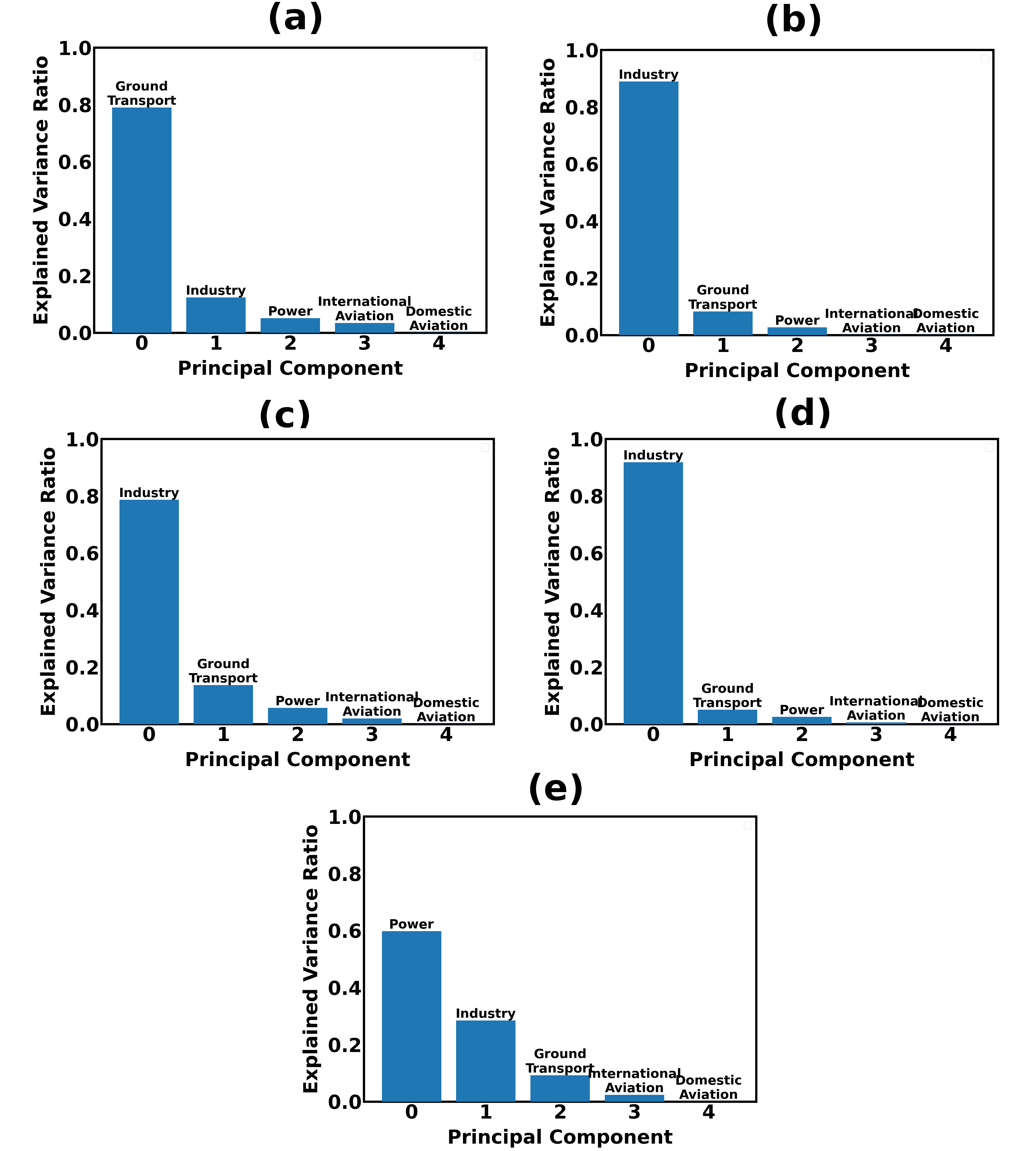}
\end{center}
\caption{Explained variance ratio for principal components of five countries, (a) EU27 \& UK, (b) India, (c) Italy, (d) Germany, and (e) Spain. The figure displays the variance ratio explained by each principal component for Domestic Aviation, Ground Transport, Industry, International Aviation, and Power sectors. The first three principal components are considered for further study and future prediction analysis.}
\label{fig5}
\end{figure}

We conduct a principal component analysis (PCA) to explore the relationship between different principal components and $CO_2$ emissions. It allows us to reduce the dimensionality of the data while retaining the most informative features for subsequent analysis and prediction tasks. Figure \ref{fig5} displays the explained variance ratio per principal component, categorized by sectors for all the countries considered. The explained variance ratio for each component indicates the proportion of the total variance in the data that is accounted for by that particular component. A value of $0$ indicates that the principal component does not explain any variance in the data, while a value of $1$ indicates that the principal component explains all of the variance. Additionally, Table \ref{tab1} presents the values of explained variance ratio for all the components for five countries. This allows us to examine the relative importance of each principal component in explaining the variability in $CO_2$ emissions for different countries. Based on the results of the PCA, we further identify the first three principal components or sectors, viz. Power, Industry, and Ground Transport as the significant contributors to further analysis and future prediction. These principal components collectively account for a substantial portion of the overall variance in the data set. By considering these three components, we can effectively capture and represent the essential patterns and trends in $CO_2$ emissions data. By utilizing the identified principal components, we can potentially develop robust models for forecasting $CO_2$ emissions and assessing the impact of various factors on emission patterns.

\begin{table}[htbp]
\centering
\caption{Explained variance ratio for principal components of five countries}
\begin{tabularx}{\textwidth}{|c|*{5}{X|}}
\hline
Country & Domestic Aviation & Ground Transport & Industry & International Aviation & Power \\
\hline
EU27 \& UK & 0.00003 & 0.78993 & 0.12394 & 0.03464 & 0.05145 \\
India & 0.000001 & 0.08291 & 0.88959 & 0.00025 & 0.02723 \\
Italy & 0.00009 & 0.13656 & 0.78689 & 0.01991 & 0.05655 \\
Germany & 0.00002 & 0.05039 & 0.91855 & 0.00574 & 0.02530 \\
Spain & 0.00047 & 0.09260 & 0.28488 & 0.02388 & 0.59817 \\
\hline
\end{tabularx}
\label{tab1}
\end{table}
\subsection{Machine Learning Analysis}
\begin{figure}[hbtp]
\begin{center}

\includegraphics[width=9cm]{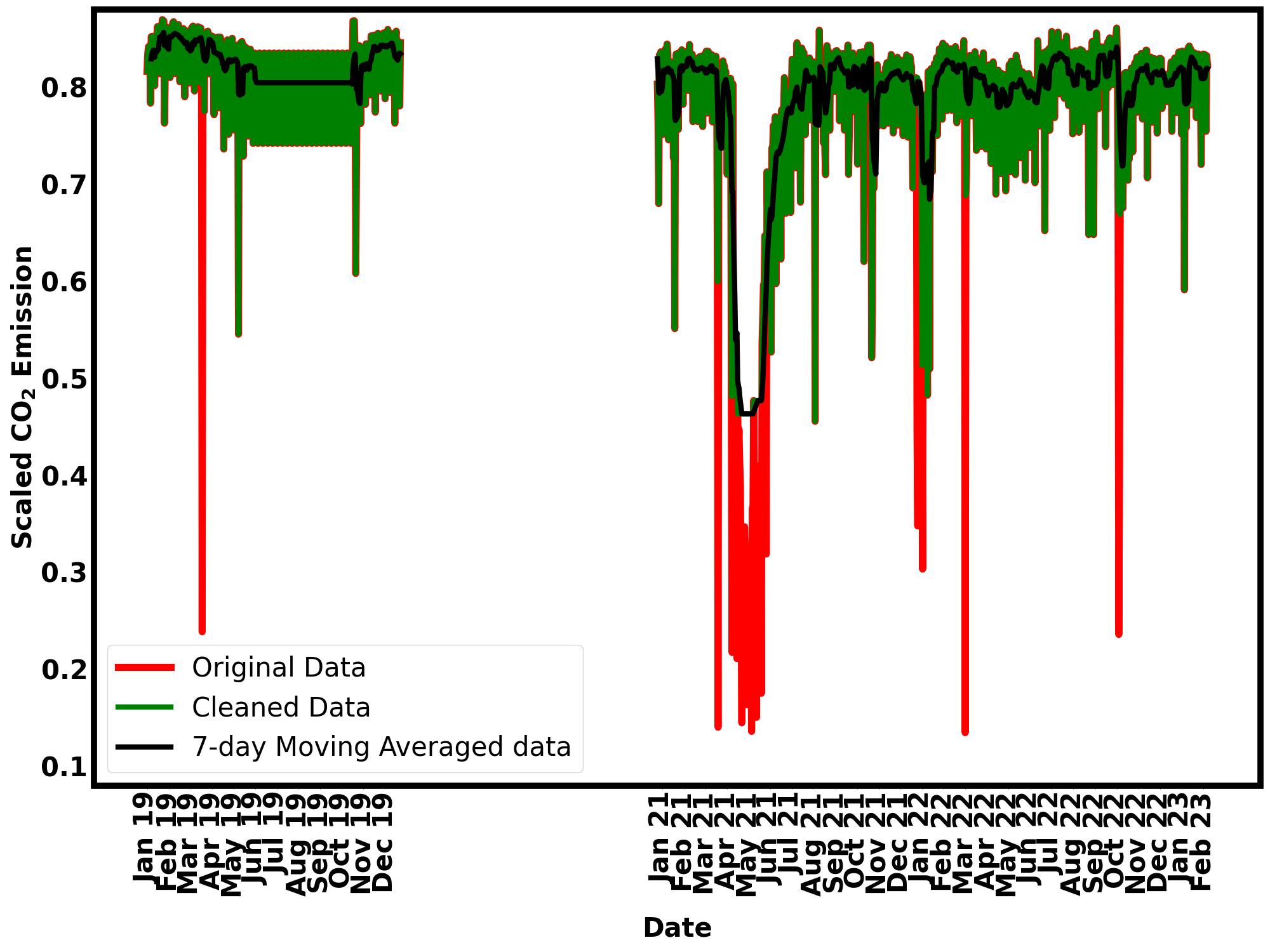}
\end{center}
\caption{Comparison of the original dataset, the cleaned dataset after outlier removal, and the 7-day moving averaged dataset. Note: $2020$ data is excluded from the analysis, which is justified due to the anomalous and outlier effects caused by the COVID-$19$ pandemic, ensuring more reliable and accurate predictions based on the regular and predictable emission patterns observed in other years.}
\label{fig6}
\end{figure}

To conduct the prediction analysis,  we first preprocess the data by removing outliers from the original dataset. The employed method, Z-score, is a statistical measure that quantifies how many standard deviations a data point is from the mean of a distribution \cite{cousineau2010outliers, tripathy2013comparison, sandbhor2019impact}. Z-score is computed by $Z=(x_i-\langle x \rangle)/SD$ where $x_i$ is the $i$i-th data-point, $\langle x \rangle$ is the mean of the dataset with standard deviation SD. We consider all the data-points greater than the threshold value $3$ as outliers. By using the Z-score method, we can identify and remove data points that deviate significantly from the average. Any data points that fall outside this threshold are considered outliers and subsequently replaced with the previous day's value, resulting in a cleaned dataset. For the prediction analysis, we utilize a moving average of the data of window size $=7$ days. This moving average smooths out short-term fluctuations and provides a clearer trend over time. By considering the average value over a $7$-day period, we can better capture the overall pattern and reduce the impact of daily variations. Figure \ref{fig6} illustrates the comparison of the three datasets: the original dataset, the cleaned dataset after outlier removal, and the $7$-day moving averaged dataset. This visualization allows for a visual assessment of the effects of the outlier removal and the smoothing effect of the moving average. 

The choice of a moving average for prediction analysis is motivated by its ability to reduce noise and reveal underlying trends in the data. This can result in more accurate and stable predictions by providing a smoother representation of the data's overall behavior. If the window size is too small, it may not adequately capture the underlying trend and instead emphasize noise or daily variations. On the other hand, too large window size will oversmooth the data, obscuring important variations and changes and become slower to respond to shifts or fluctuations in the data. By focusing on a $7$-day period, we can capture trends that occur over a few weeks or months without sacrificing too much detail and maintain a balance between capturing long-term trends and remaining responsive to recent changes in the data. These steps aim to enhance the quality and reliability of the data, facilitating more robust predictions and a better understanding of the underlying trends.

\begin{figure}[hbtp]
\begin{center}
\includegraphics[max height=1.2\textwidth,max width=\textwidth, keepaspectratio]{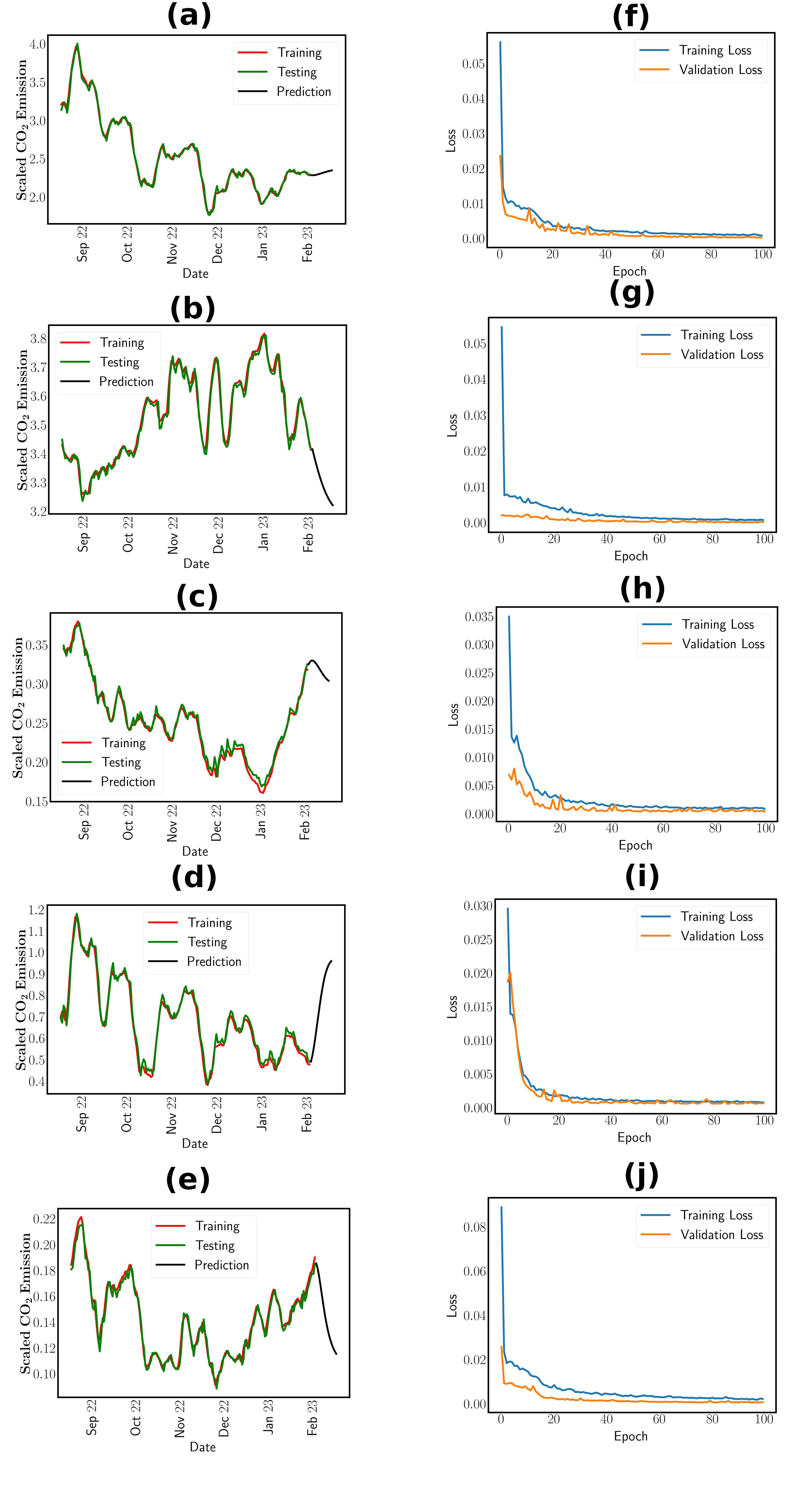}
\end{center}
\caption{(Left)Comparison of Training, Testing, and Prediction Results, (Right) Loss Function Plot during LSTM Model Training and validation of Power sector Emissions for EU27 \& UK, India, Italy, Germany, and Spain respectively (from top to bottom).}
\label{fig7}
\end{figure}

\begin{figure}[hbtp]
\begin{center}
\includegraphics[max height=1.2\textwidth,max width=\textwidth, keepaspectratio]{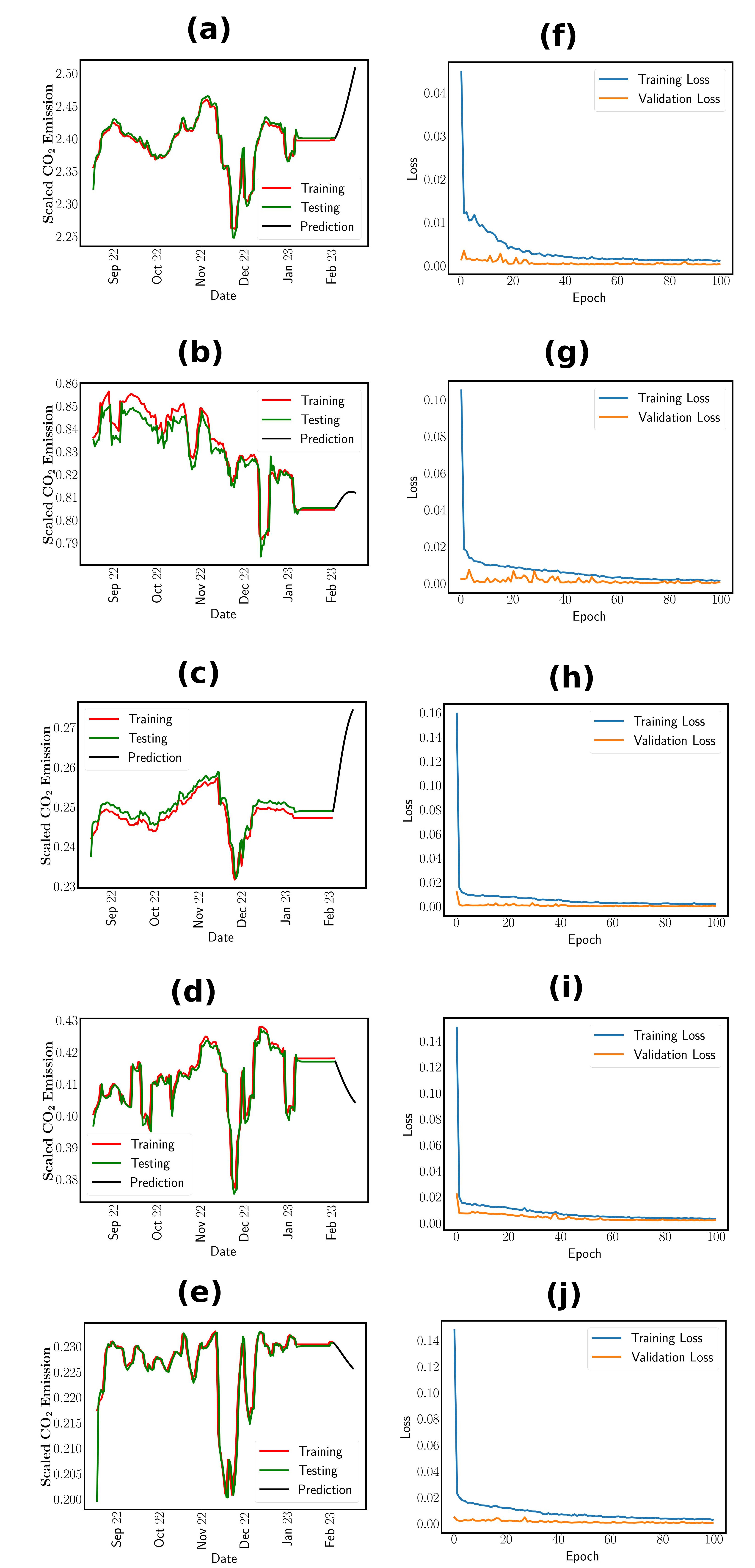}
\end{center}
\caption{(Left)Comparison of Training, Testing, and Prediction Results, (Right) Loss Function Plot during LSTM Model Training and validation of Ground Transport sector Emissions for EU27 \& UK, India, Italy, Germany, and Spain, respectively (from top to bottom).}
\label{fig8}
\end{figure}

\begin{figure}[hbtp]
\begin{center}
\includegraphics[max height=1.2\textwidth,max width=\textwidth, keepaspectratio]{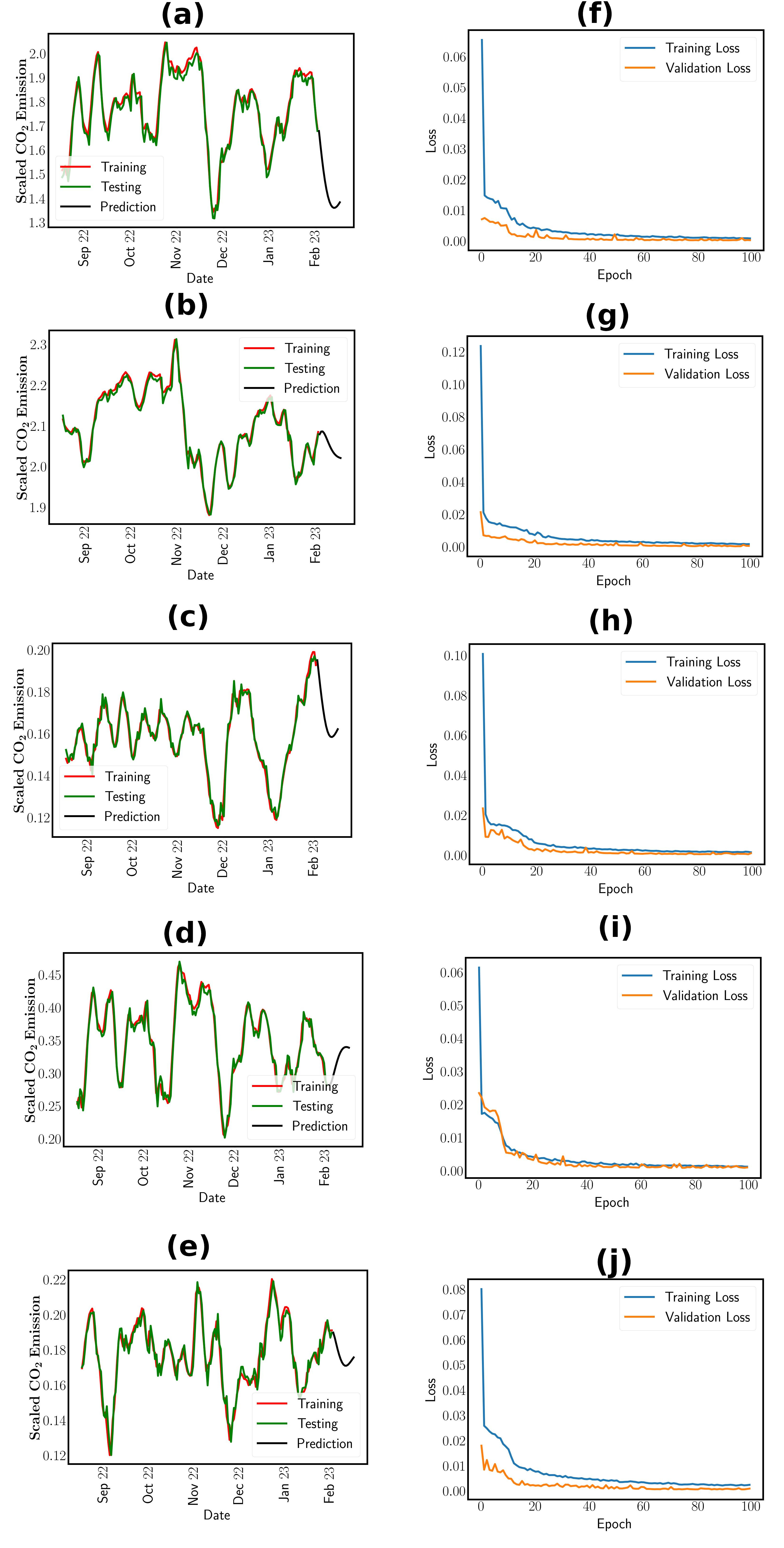}
\end{center}
\caption{(Left)Comparison of Training, Testing, and Prediction Results, (Right) Loss Function Plot during LSTM Model Training and validation of Industry sector Emissions for EU27 \& UK, India, Italy, Germany, and Spain, respectively (from top to bottom).}
\label{fig9}
\end{figure}

The analysis of the results presented in Fig. \ref{fig7}, \ref{fig8}, and \ref{fig9}, for Power, Ground Transport, and Industry sectors, respectively, demonstrate the effectiveness of the LSTM model (Left column) in capturing patterns and trends in the time series data. To optimize the model's performance for different datasets, parameter tuning is required. We conducted experiments to identify the best results by exploring batch sizes in the range of $8$ to $32$, dropout values ranging from $0.1$ to $0.3$, and training the model for $100$ epochs (refer to Table \ref{tab2} for the details). The visualizations indicate that the model has accurately predicted the values. Furthermore, the evaluation of the loss function (Right column of Fig. \ref{fig7}, \ref{fig8}, and \ref{fig9}) during model training reveals that the model has achieved convergence and remained stable, ensuring its reliability in the testing phase. The detailed performance metrics in Table \ref{tab2}, including the lower RMSE and highly accurate MAE values, further support the model's predictive accuracy. The higher $R^2$ value, close to $1$, signifies that the model explains a significant portion of the variance in the data, indicating its strong performance.

\begin{table}[hbtp]
\caption{Performance Metrics of the LSTM Model on Test Set}
\resizebox{\textwidth}{!}{%
\begin{tabular}{|l|l|l|l|l|l|l|l|}
\hline
\textbf{Country} & \textbf{Evaluating parameters} & \textbf{Power} & \textbf{} & \textbf{Industry} & \textbf{} & \textbf{Ground Transport} & \textbf{} \\ \hline
\textbf{EU27 \& UK} & MSE & 0.0013 & Batch size- 32 & 0.0006 & Batch size- 16 & 0.0001 & Batch size- 8 \\ \hline
\textbf{} & RMSE & 0.036 & Dropout- 0.16 & 0.025 & Dropout- 0.16 & 0.0112 & Dropout- 0.2 \\ \hline
\textbf{} & MAE & 0.0277 & Epochs- 100 & 0.019 & Epochs- 100 & 0.0059 & Epochs- 100 \\ \hline
\textbf{} & \textbf{$R^2$} & \textbf{0.995} &  & \textbf{0.9734} &  & \textbf{0.9106} &  \\ \hline
\textbf{India} & MSE & 0.0003 & Batch size- 8 & 0.0001 & Batch size- 16 & 2E-05 & Batch size- 32 \\ \hline
\textbf{} & RMSE & 0.0173 & Dropout- 0.1 & 0.0109 & Dropout- 0.1 & 0.0046 & Dropout- 0.16 \\ \hline
\textbf{} & MAE & 0.0135 & Epochs- 100 & 0.008 & Epochs- 100 & 0.0031 & Epochs- 100 \\ \hline
\textbf{} & \textbf{$R^2$} & \textbf{0.9859} &  & \textbf{0.9859} &  & \textbf{0.9353} &  \\ \hline
\textbf{Italy} & MSE & 2.1E-05 & Batch size- 16 & 8.92E-06 & Batch size- 32 & 1.6E-06 & Batch size- 16 \\ \hline
\textbf{} & RMSE & 0.0047 & Dropout- 0.16 & 0.003 & Dropout- 0.16 & 0.0013 & Dropout- 0.3 \\ \hline
\textbf{} & MAE & 0.0036 & Epochs- 100 & 0.0023 & Epochs- 100 & 0.0006 & Epochs- 100 \\ \hline
\textbf{} & \textbf{$R^2$} & \textbf{0.9917} &  & \textbf{0.9711} &  & \textbf{0.9188} &  \\ \hline
\textbf{Germany} & MSE & 0.0004 & Batch size- 8 & 0.0001 & Batch size- 32 & 5.9E-06 & Batch size- 32 \\ \hline
\textbf{} & RMSE & 0.0207 & Dropout- 0.2 & 0.0103 & Dropout- 0.1 & 0.0024 & Dropout- 0.1 \\ \hline
\textbf{} & MAE & 0.0157 & Epochs- 100 & 0.0081 & Epochs- 100 & 0.0013 & Epochs- 100 \\ \hline
\textbf{} & \textbf{$R^2$} & \textbf{0.9873} &  & \textbf{0.9701} &  & \textbf{0.8841} &  \\ \hline
\textbf{Spain} & MSE & 1.45E-05 & Batch size- 32 & 2.1E-05 & Batch size- 32 & 1.6E-06 & Batch size- 32 \\ \hline
\textbf{} & RMSE & 0.0038 & Dropout- 0.3 & 0.0046 & Dropout- 0.3 & 0.004 & Dropout- 0.3 \\ \hline
\textbf{} & MAE & 0.0031 & Epochs- 100 & 0.0036 & Epochs- 100 & 0.002 & Epochs- 100 \\ \hline
\textbf{} & \textbf{$R^2$} & \textbf{0.9834} &  & \textbf{0.9452} &  & \textbf{0.8242} &  \\ \hline
\end{tabular}%
}
\label{tab2}
\end{table}
\subsection{DFT Calculations}
In the context of capturing gases from the environment, ensuring the stability of designed materials is crucial for their practical applications. Before utilizing these materials for $CO_2$ capture or activation, it is imperative to investigate their stability thoroughly. To address this, we have developed a novel category of two-dimensional materials that exhibit promising adsorption and catalytic activity concerning flue gas molecules.
Our proposal involves using aluminium-doped MBenes, which are more efficient in $CO_2$ activation compared to standard MBenes\cite{VKY0}. The MBenes we considered consist of two transition metal atoms and two boron or aluminium atoms within a single unit cell, with each atom connected to six neighbouring atoms. This contrasts MXenes, where the surface metal atoms are coordinated to three carbon or nitrogen atoms.
\subsubsection{Structural and Electronic Properties}
\begin{figure}[t]
\begin{center}
\includegraphics[width=10cm]{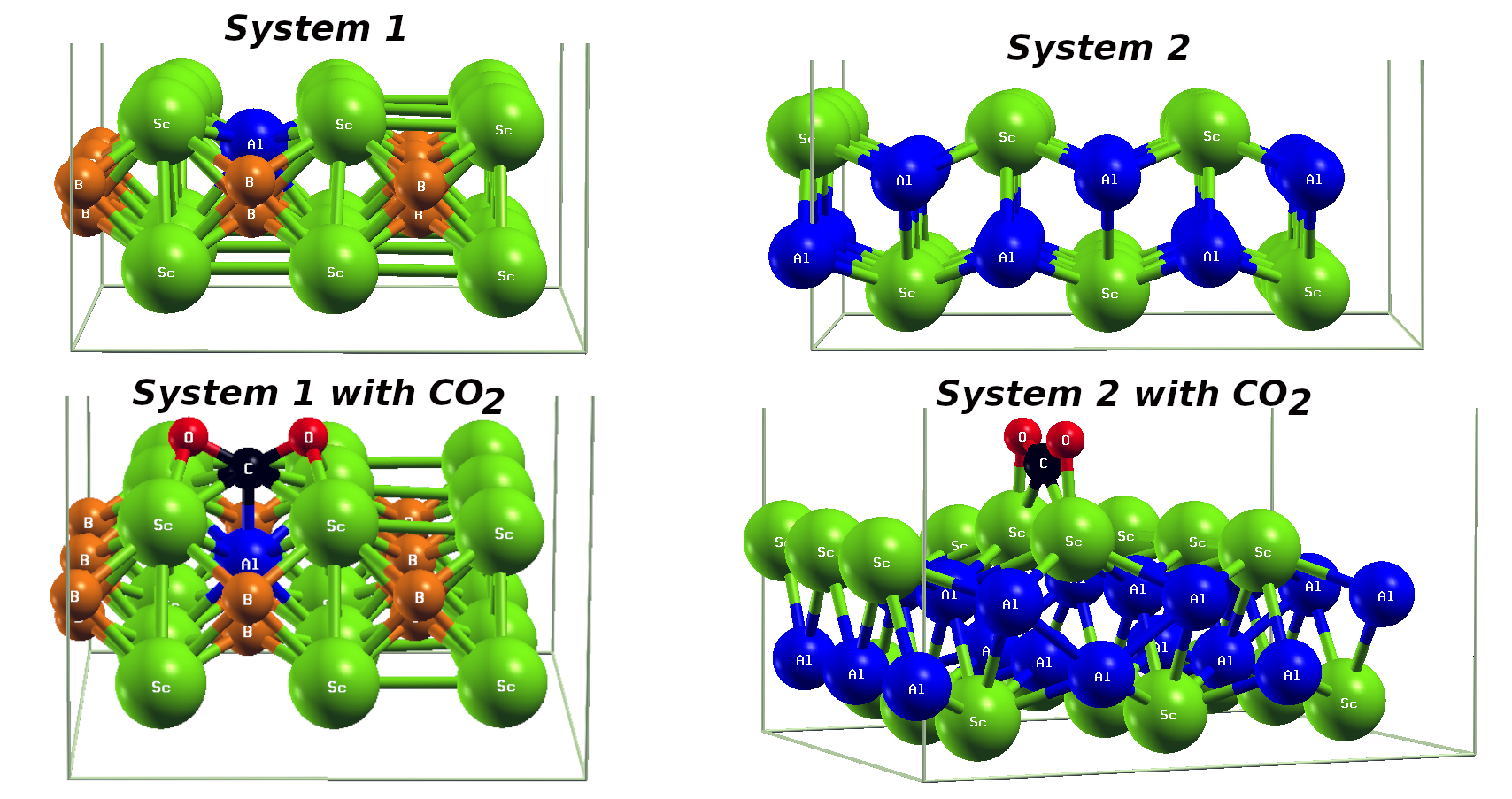}
\end{center}
\caption{Top and side views of systems 1 and 2 schematic structure, without and with $CO_2$ adsorption, respectively.}
\label{fig10}
\end{figure}

\begin{figure}[t]
\begin{center}
\includegraphics[width=10cm]{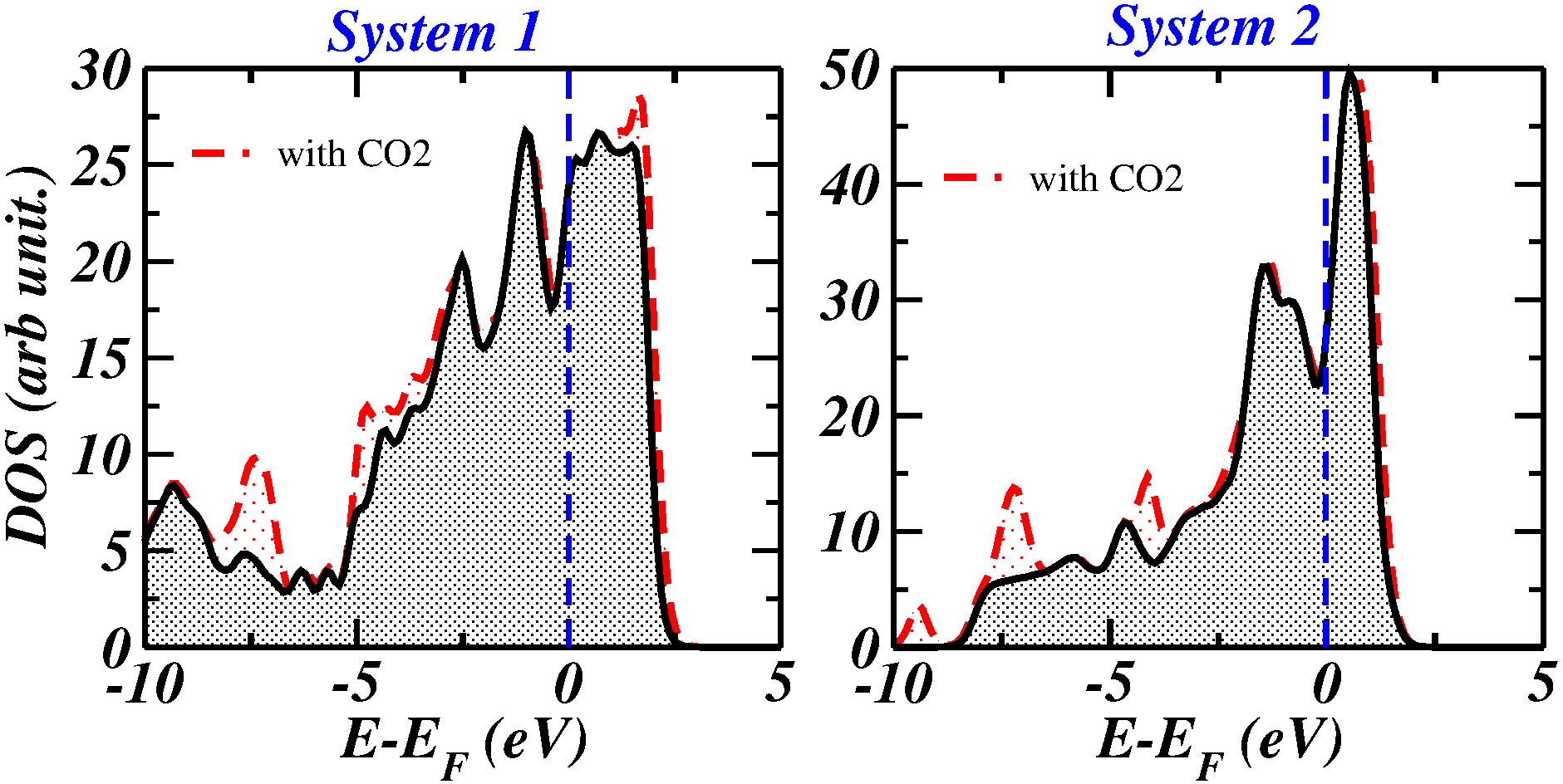}
\end{center}
\caption{Density of States (DOS) for systems $1$ and $2$, with and without $CO_2$, respectively.}
\label{fig11}
\end{figure}

\begin{figure}[t]
\begin{center}
\includegraphics[width=10cm]{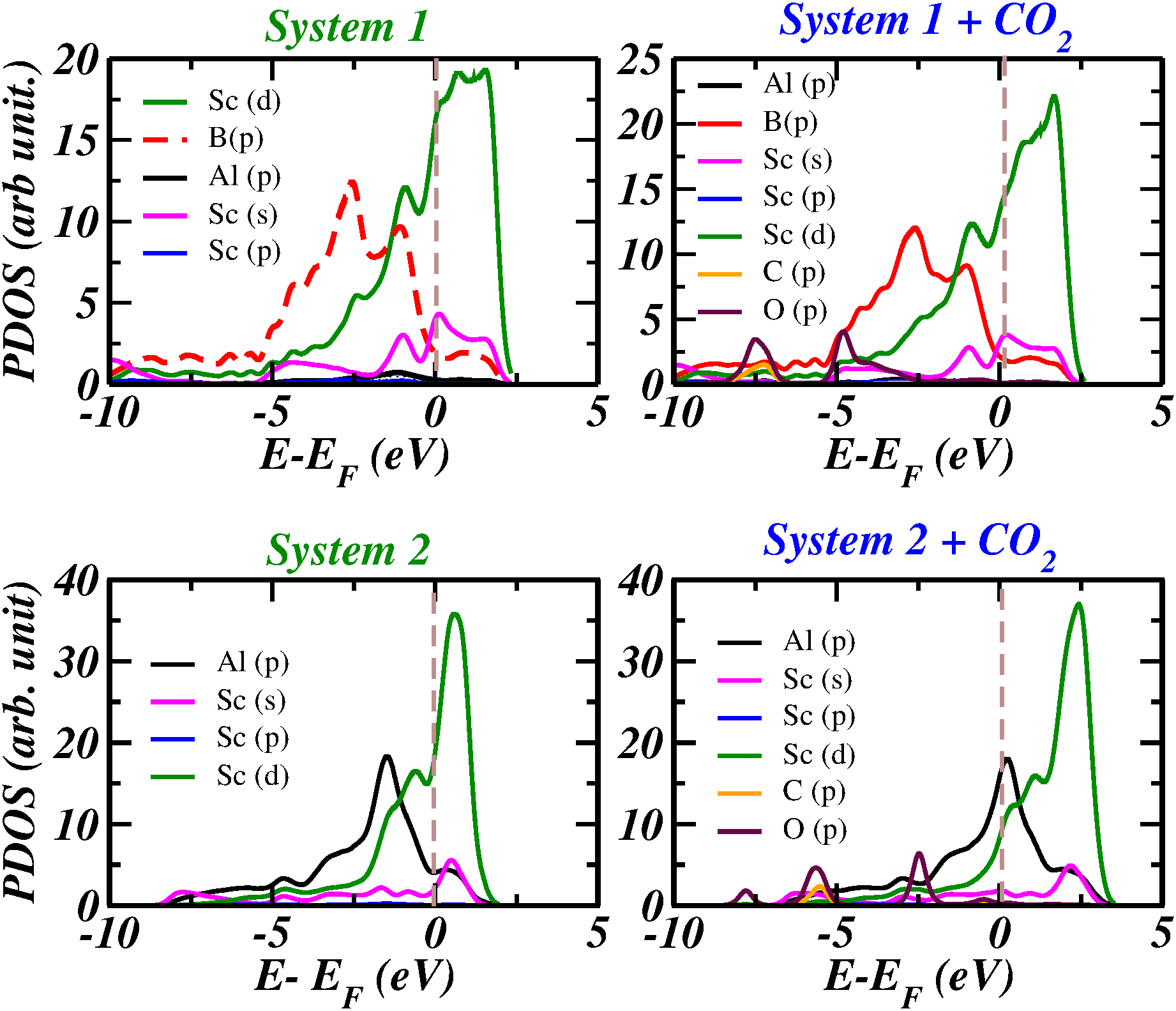}
\end{center}
\caption{Projected Density of States (PDOS) for systems $1$ and $2$, with and without $CO_2$, respectively.}
\label{fig12}
\end{figure}

\begin{figure}[t]
\begin{center}
\includegraphics[width=10cm]{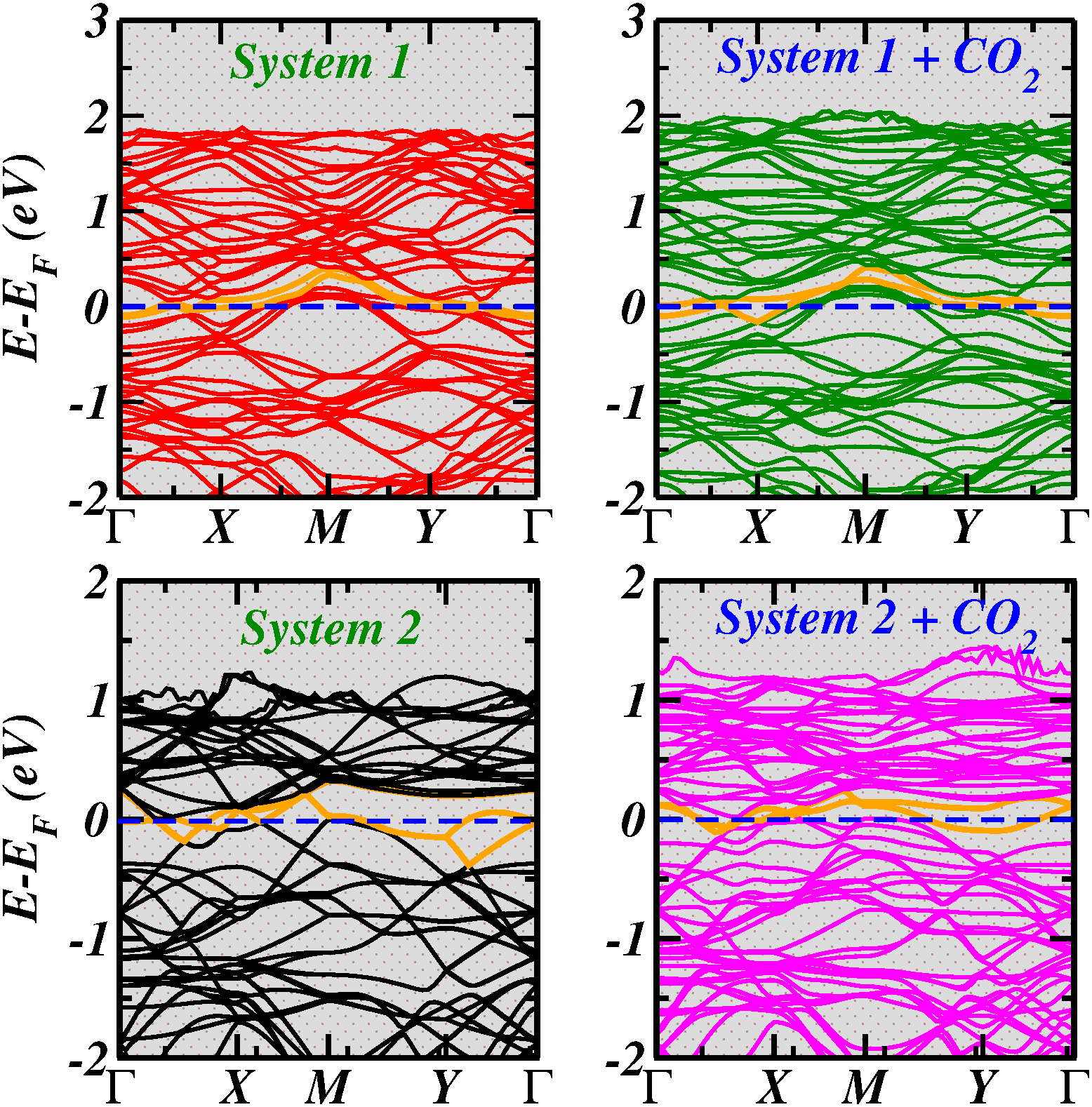}
\end{center}
\caption{Bands Structure for systems $1$ and $2$, with and without $CO_2$, respectively.}
\label{fig13}
\end{figure}

The structure of designed two-dimensional materials with different doping concentrations of aluminium is shown in Fig. \ref{fig10}. The structure Fig. \ref{fig10}(a) contains one boron atom replaced by an aluminium atom, and (b) all boron is replaced by aluminium atoms in Scandium-based MBenes. These structures exhibit Pmma (no.51) type space group, which results in in-plane structural anisotropy in their mechanical properties. Like MBenes, these structures also have metal exposed surfaces with buckled bilayers. The computed distances between the M = Scandium and boron and aluminium are given in Table \ref{tab3}. This distance is the direct covalent bond between the B/Al with the scandium atom. After the adsorption of $CO_2$ over system $1$ and system $2$, it was found that in the presence of a single aluminium atom (system $1$), the $CO_2$ goes deep inside the cavity around the aluminium atom (as shown in Fig. \ref{fig10} (a) bottom). Due to this, it shows stronger interaction with system $1$. For the case of system $2$, due to the uniform distribution of aluminium into the sheet, the $CO_2$ prefers to be on the bridge side of exposed scandium metal and makes bonds with only scandium metal. The possible reason for this behaviour of $CO_2$ is probably due to metal to non-metal ratio. System $1$ contains metal and non-metal, whereas System $2$ has only metals in the sheet. This stronger interaction of $CO_2$ with system $1$ leads towards more negative interaction energy than system $2$, as shown in table \ref{tab3}.
For activation of $CO_2$ and conversion into useful production, the prerequisite is to make $CO_2$ bend from its linear form and elongate the CO bond length. The $CO_2$ when adsorbed over system $1$ and system $2$, $CO_2$ transform from linear to a non-linear moiety, and their CO bond length also increases. This change in bond length and bond angle of $CO_2$ from its stable gas phase makes it active for further conversion over the sheet. 
\\
For the adsorption of $CO_2$ over system $1$, the bond length of ($d_{CO}$ 
increases from $1.17$ to $1.37$ \AA) whereas the $CO_2$ bond angle 
($\angle$(OCO) reduces from $180$ to $137$$\degree$). Similar behaviour is 
also observed for the $CO_2$ adsorption over system $2$; here, the bond length 
of ($d_{CO}$ increases from $1.17$ to $1.28$ \AA) and $CO_2$ bond angle 
($\angle$(OCO) also reduces from $180$ to $131$$\degree$). The difference in $CO_2$ bond length and angle over system $1$ and system $2$ can be justified by how it interacts with the sheet and the local electronic arrangements around it and $CO_2$. Similarly, the distance between two oxygen atoms ($d_{OO}$), aluminium-carbon ($d_{AlC}$), and scandium-scandium ($d_{ScSc}$) for system $1$ (system $2$) have been computed, and their values are found to be $2.41$ ($2.32$), $1.94$ ($3.41$), and $3.40$ ($3.07$) \AA, respectively.  
\\
To investigate the electronic properties of systems $1$ and $2$, we have computed the density of state (DOS), projected density of states (PDOS) and band structure by employing the traditional PBE functional as shown in Fig. \ref{fig11}, Fig. \ref{fig12} and Fig. \ref{fig13} respectively. The DOS figure (Fig. \ref{fig11}) for both system $1$ and system $2$ clearly shows the density around the Fermi energy level, which indicates that the designed systems are metallic in nature with nearly zero band gap. This gap-less behaviour continues after the adsorption of $CO_2$ over the sheets as the HOMO and LUMO of the $CO_2$ do not participate towards the bandgap. This also indicates that the adsorption doesn't affect the property of sheets and only alters the local environment. 

To elucidate the origin of the high adsorption energy of $CO_2$, we also computed the PDOS of $CO_2$ adsorbed over both systems and compared it with the PDOS of pristine systems. It is clear from the PDOS for system $1$ that the contribution from the p-orbital of the metal and boron/aluminium is less reactive, which signifies no DOS from the p-orbital of the metal at the Fermi level. The main contribution arises only from the d-orbital of the metal at the Fermi level. For system $2$, the PDOS behave similarly to system $1$ (without $CO_2$) and is dominated by the d-orbital contribution of scandium metal towards the Fermi level. In the $CO_2$ adsorbed system $2$, the electronic arrangement pushes the p-orbital of aluminium towards the Fermi level and the d-orbital of scandium metal. Overall, we can conclude that the systems remain metallic before and after adsorption.

Figure \ref{fig13} also confirms the metallic nature when expanding in the desired Brillouin zone for band structure calculations. The bands cross the Fermi level (blue dashed line at 0) for both systems before and after the adsorption of $CO_2$, as marked by the yellow line in Fig. \ref{fig13}.   
\subsubsection{\texorpdfstring{$CO_2$}{CO2} Adsorption Analysis}
The cohesive/formation energy is calculated using the equation mentioned in the methodology section as eqn. \ref{eqn6}. We computed the energy of the bare atoms involved in forming the materials and subtracted them with the energy of the 2d sheet. The formation energy (from eqn. \ref{eqn7}) of system $1$ and system $2$ is found to be -5.81 and -4.07 eV, respectively, which is comparable with the Mbenes ($Sc_2B_2$), whose formation energy is found to be -$5.71$ eV \cite{VKY8}. This indicates that our designed two-dimensional sheets are stable. We also conclude that systems $1$ and $2$ have a negative formation energy, indicating their stability relative to the bulk constituents. We found that our proposed $2D$ sheets show good stability and can be verified using phonon calculation in the future.\\
The strong interaction of the $CO_2$ with sheets results in the chemical bond formation between the O(C) atom of $CO_2$ and the sheet. This results in strong interaction and hence large adsorption energy of $CO_2$ on system $1$ and system $2$. The adsorption/binding energy of $CO_2$ on system $1$ and system $2$ are -3.31 and -2.92, in eV, respectively (see Table \ref{tab3} for details).

\begin{table}[htbp]
\centering
\caption{Results of DFT Calculations}
\begin{tabularx}{\textwidth}{|>{\centering\arraybackslash}p{5cm}|*{5}{>{\centering\arraybackslash}X|}}
\hline
\multicolumn{1}{|c|}{\multirow{2}{*}{Properties}} & \multicolumn{2}{c|}{Designed Materials} \\ \cline{2-3} 
\multicolumn{1}{|c|}{} & \multicolumn{1}{c|}{system $1$($Sc_{18}Al_{1}B_{17}$)} & \multicolumn{1}{c|}{system $2$ ($Sc_{18}Al_{18}$)}  \\ \hline
Lattice parameter (a) in \AA & 9.3738 & 12.7114 \\ \hline
Lattice parameter (b) in \AA & 9.9669 & 9.8719 \\ \hline
Cohesive/Formation Energy (eV) & -5.81 & -4.07 \\ \hline
HOMO(with $CO_2$) in eV & 1.9168(2.0538) & 1.6730(1.6284) \\ \hline
LUMO(with $CO_2$) in eV & 1.9186(2.0620) & 1.6755(1.6410)  \\ \hline
$E_{fermi}$ (with $CO_2$) in eV& 1.9176(2.0601) & 1.6748(1.6330) \\ \hline
Band Gap (with $CO_2$) in eV & 0.0018(0.0082) & 0.0025(0.0126) \\ \hline
$CO_2$ Binding Energy (eV) & -3.31 & -2.92 \\ \hline
$d_{CO}$ \AA& 1.37 & 1.28 \\ \hline
$\angle$(OCO) $\degree$& 137 & 131  \\ \hline
\end{tabularx}
\label{tab3}
\end{table}

Finally, we constructed both systems' HOMO and LUMO plots to elucidate the electronic arrangement. In both systems, the red and blue mesh shows the HOMO and LOMO, respectively, with the isosurface value of $0.009$ $e/\mathring{A}^{3}$. It is concluded that charge accumulation or depletion occurs mostly around the $CO_2$ molecule when it interacts with system $1$ or system $2$, which may be the possible cause for its strong interaction with the surface of the proposed sheets.\\ 
\begin{figure}[t]
\begin{center}
\includegraphics[width=10cm]{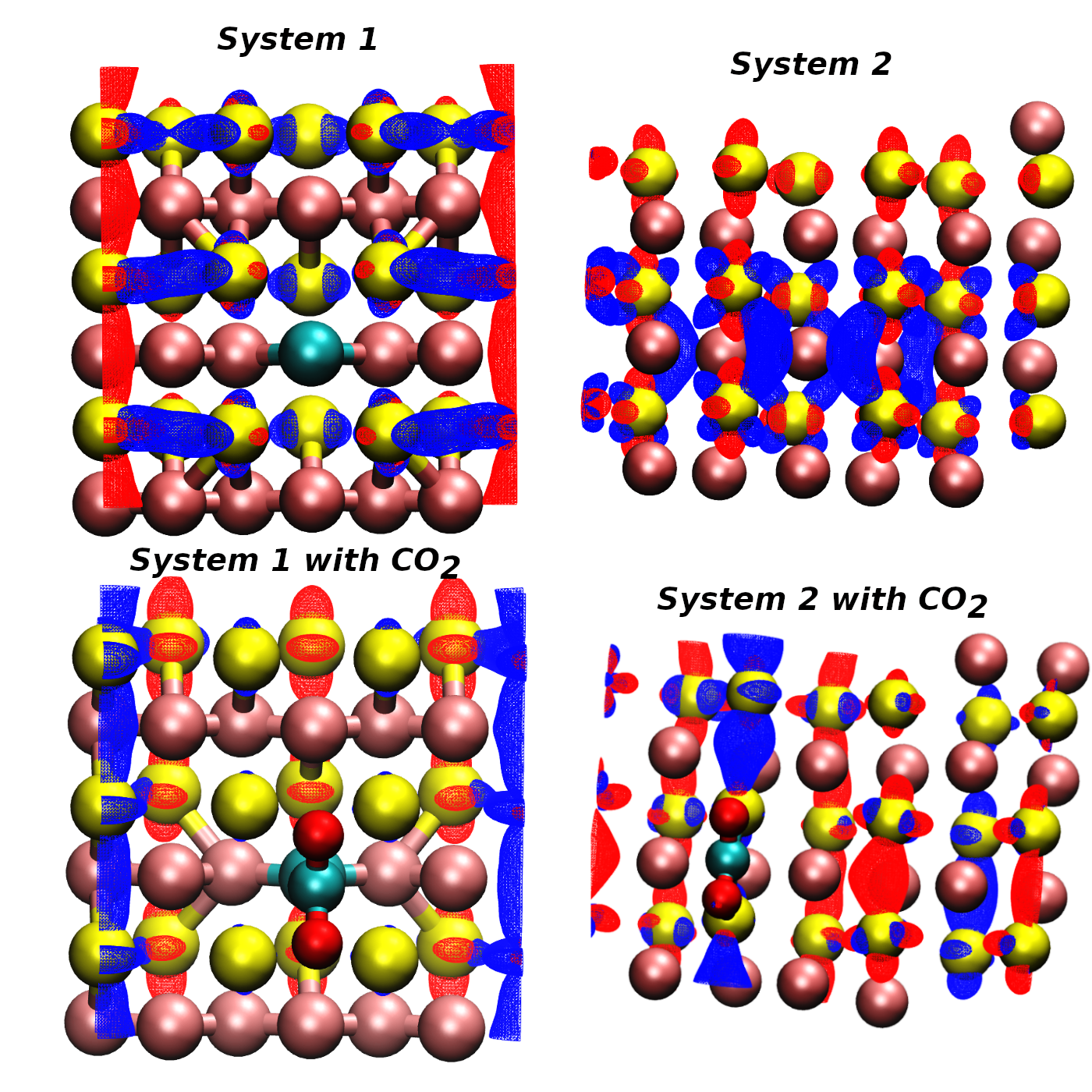}
\end{center}
\caption{HOMO(red) and LUMO(blue) plots for systems $1$ and $2$, with and without $CO_2$, respectively. The isosurface value is set at $0.009$ $e/\mathring{A}^{3}$}
\label{fig14}
\end{figure}
Similarly, we also proposed and demonstrated the adsorption and activation of $CO_2$ for the System 3 $(Sc_{18}Al_{9}B_{9})$ sheet, where the two $CO_2$ molecules can be captured and activated from both sides. Fig. \ref{fig15} shows that the Boron exposed phase attracted the $CO_2$ faster than the Aluminium exposed face, which can be concluded from the larger lattice arrangement required for aluminium compared to the boron phase. 
Boron and Aluminium belong to the same group but have different sizes, so faster capturing is preferred on the Boron side compared to the Aluminium side. The GIF movie demonstrating this dual-side adsorption is shown in SI1.
\begin{figure}[t]
\begin{center}
\includegraphics[width=8cm]{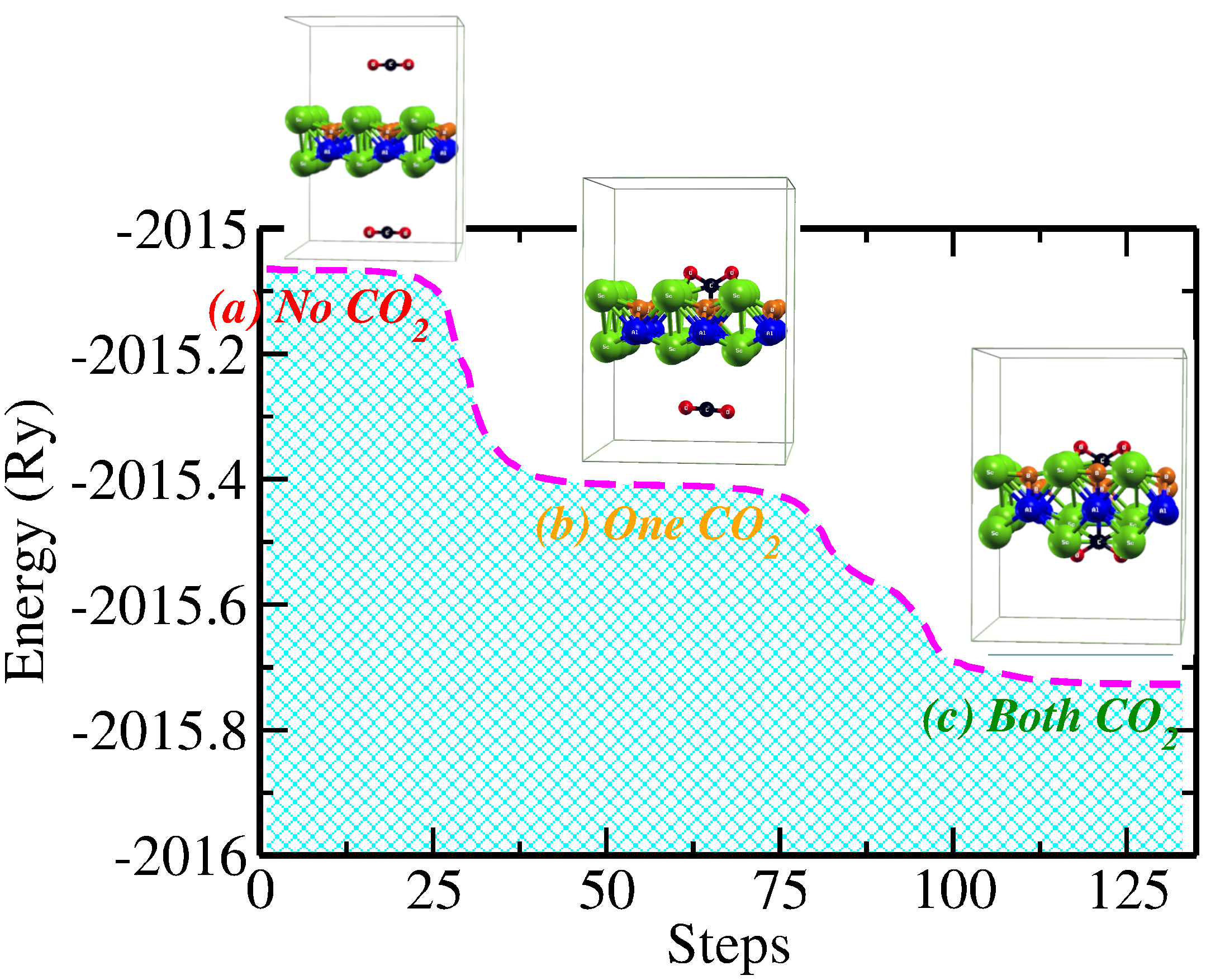}
\end{center}
\caption{Adsorption energy of $CO_2$ from both sides of sheets.}
\label{fig15}
\end{figure}

\section{Conclusion}\label{con}
This study comprehensively analyzed $CO_2$ emissions before and after the COVID-$19$ pandemic to understand its impact and predict emission patterns. The study revealed disparities in emissions between the pandemic year $(2020)$ and other years, indicating the influence of unique circumstances such as economic disruptions and changes in energy consumption patterns. To focus on regular emission patterns, the exclusion of $2020$ data was implemented, enabling a clearer understanding of underlying dynamics and more accurate predictions of emission patterns.
We identified the Power, Industry, and Ground Transport sectors as significant contributors to emission patterns for the considered countries, collectively explaining a substantial portion of the dataset's variance. These findings pave the way for the development of robust models to forecast $CO_2$ emissions and assess the impact of various factors.
Our LSTM models effectively captured patterns and trends in $CO_2$ emissions for specific sectors, demonstrating accurate predictions and high predictive accuracy. These models provide valuable insights for policy decisions, mitigation strategies, and climate change mitigation efforts, emphasizing the importance of considering sector-specific factors and the impact of unique circumstances like the pandemic.
From a material design perspective, higher environmental $CO_2$ concentration can be reduced by preparing thin films that efficiently capture and activate the $CO_2$ into valuable chemicals. The current work shows excellent stability and adsorbs $CO_2$ with higher affinity ($-3.3$ eV with system $1$ and $-2.9$ eV with system $2$) compared with graphene, boron nitride sheets and other similar two-dimensional materials.
\bibliography{bibliography}

\providecommand{\latin}[1]{#1}
\makeatletter
\providecommand{\doi}
  {\begingroup\let\do\@makeother\dospecials
  \catcode`\{=1 \catcode`\}=2 \doi@aux}
\providecommand{\doi@aux}[1]{\endgroup\texttt{#1}}
\makeatother
\providecommand*\mcitethebibliography{\thebibliography}
\csname @ifundefined\endcsname{endmcitethebibliography}
  {\let\endmcitethebibliography\endthebibliography}{}
\begin{mcitethebibliography}{61}
\providecommand*\natexlab[1]{#1}
\providecommand*\mciteSetBstSublistMode[1]{}
\providecommand*\mciteSetBstMaxWidthForm[2]{}
\providecommand*\mciteBstWouldAddEndPuncttrue
  {\def\EndOfBibitem{\unskip.}}
\providecommand*\mciteBstWouldAddEndPunctfalse
  {\let\EndOfBibitem\relax}
\providecommand*\mciteSetBstMidEndSepPunct[3]{}
\providecommand*\mciteSetBstSublistLabelBeginEnd[3]{}
\providecommand*\EndOfBibitem{}
\mciteSetBstSublistMode{f}
\mciteSetBstMaxWidthForm{subitem}{(\alph{mcitesubitemcount})}
\mciteSetBstSublistLabelBeginEnd
  {\mcitemaxwidthsubitemform\space}
  {\relax}
  {\relax}

\bibitem[{United Nations, Department of Economic and Social Affairs, Population
  Division}(2022)]{UNPopulation2022}
{United Nations, Department of Economic and Social Affairs, Population
  Division}, \emph{World Population Prospects 2022: Summary of Results};
  2022\relax
\mciteBstWouldAddEndPuncttrue
\mciteSetBstMidEndSepPunct{\mcitedefaultmidpunct}
{\mcitedefaultendpunct}{\mcitedefaultseppunct}\relax
\EndOfBibitem
\bibitem[Bonga and Chirowa(2014)Bonga, and Chirowa]{bonga2014level}
Bonga,~W.~G.; Chirowa,~F. Level of cooperativeness of individuals to issues of
  energy conservation. \emph{Available at SSRN 2412639} \textbf{2014}, \relax
\mciteBstWouldAddEndPunctfalse
\mciteSetBstMidEndSepPunct{\mcitedefaultmidpunct}
{}{\mcitedefaultseppunct}\relax
\EndOfBibitem
\bibitem[Olivier \latin{et~al.}(2017)Olivier, Schure, Peters, \latin{et~al.}
  others]{olivier2017trends}
Olivier,~J.~G.; Schure,~K.; Peters,~J., \latin{et~al.}  Trends in global CO2
  and total greenhouse gas emissions. \emph{PBL Netherlands Environmental
  Assessment Agency} \textbf{2017}, \emph{5}, 1--11\relax
\mciteBstWouldAddEndPuncttrue
\mciteSetBstMidEndSepPunct{\mcitedefaultmidpunct}
{\mcitedefaultendpunct}{\mcitedefaultseppunct}\relax
\EndOfBibitem
\bibitem[A{\u{g}}bulut(2022)]{augbulut2022forecasting}
A{\u{g}}bulut,~{\"U}. Forecasting of transportation-related energy demand and
  CO2 emissions in Turkey with different machine learning algorithms.
  \emph{Sustainable Production and Consumption} \textbf{2022}, \emph{29},
  141--157\relax
\mciteBstWouldAddEndPuncttrue
\mciteSetBstMidEndSepPunct{\mcitedefaultmidpunct}
{\mcitedefaultendpunct}{\mcitedefaultseppunct}\relax
\EndOfBibitem
\bibitem[Bakay and A{\u{g}}bulut(2021)Bakay, and
  A{\u{g}}bulut]{bakay2021electricity}
Bakay,~M.~S.; A{\u{g}}bulut,~{\"U}. Electricity production based forecasting of
  greenhouse gas emissions in Turkey with deep learning, support vector machine
  and artificial neural network algorithms. \emph{Journal of Cleaner
  Production} \textbf{2021}, \emph{285}, 125324\relax
\mciteBstWouldAddEndPuncttrue
\mciteSetBstMidEndSepPunct{\mcitedefaultmidpunct}
{\mcitedefaultendpunct}{\mcitedefaultseppunct}\relax
\EndOfBibitem
\bibitem[Ahmadi(2019)]{ahmadi2019environmental}
Ahmadi,~P. Environmental impacts and behavioral drivers of deep decarbonization
  for transportation through electric vehicles. \emph{Journal of cleaner
  production} \textbf{2019}, \emph{225}, 1209--1219\relax
\mciteBstWouldAddEndPuncttrue
\mciteSetBstMidEndSepPunct{\mcitedefaultmidpunct}
{\mcitedefaultendpunct}{\mcitedefaultseppunct}\relax
\EndOfBibitem
\bibitem[Bistline and Rai(2010)Bistline, and Rai]{bistline2010role}
Bistline,~J.~E.; Rai,~V. The role of carbon capture technologies in greenhouse
  gas emissions-reduction models: A parametric study for the US power sector.
  \emph{Energy policy} \textbf{2010}, \emph{38}, 1177--1191\relax
\mciteBstWouldAddEndPuncttrue
\mciteSetBstMidEndSepPunct{\mcitedefaultmidpunct}
{\mcitedefaultendpunct}{\mcitedefaultseppunct}\relax
\EndOfBibitem
\bibitem[Wen and Cao(2020)Wen, and Cao]{wen2020influencing}
Wen,~L.; Cao,~Y. Influencing factors analysis and forecasting of residential
  energy-related CO2 emissions utilizing optimized support vector machine.
  \emph{Journal of Cleaner Production} \textbf{2020}, \emph{250}, 119492\relax
\mciteBstWouldAddEndPuncttrue
\mciteSetBstMidEndSepPunct{\mcitedefaultmidpunct}
{\mcitedefaultendpunct}{\mcitedefaultseppunct}\relax
\EndOfBibitem
\bibitem[Wang and Ang(2018)Wang, and Ang]{wang2018assessing}
Wang,~H.; Ang,~B. Assessing the role of international trade in global CO2
  emissions: An index decomposition analysis approach. \emph{Applied Energy}
  \textbf{2018}, \emph{218}, 146--158\relax
\mciteBstWouldAddEndPuncttrue
\mciteSetBstMidEndSepPunct{\mcitedefaultmidpunct}
{\mcitedefaultendpunct}{\mcitedefaultseppunct}\relax
\EndOfBibitem
\bibitem[Donglan \latin{et~al.}(2010)Donglan, Dequn, and
  Peng]{donglan2010driving}
Donglan,~Z.; Dequn,~Z.; Peng,~Z. Driving forces of residential CO2 emissions in
  urban and rural China: An index decomposition analysis. \emph{Energy policy}
  \textbf{2010}, \emph{38}, 3377--3383\relax
\mciteBstWouldAddEndPuncttrue
\mciteSetBstMidEndSepPunct{\mcitedefaultmidpunct}
{\mcitedefaultendpunct}{\mcitedefaultseppunct}\relax
\EndOfBibitem
\bibitem[Liu \latin{et~al.}(2019)Liu, Yang, Zhang, Sun, and
  Xu]{liu2019analysis}
Liu,~J.; Yang,~Q.; Zhang,~Y.; Sun,~W.; Xu,~Y. Analysis of CO2 emissions in
  China’s manufacturing industry based on extended logarithmic mean division
  index decomposition. \emph{Sustainability} \textbf{2019}, \emph{11},
  226\relax
\mciteBstWouldAddEndPuncttrue
\mciteSetBstMidEndSepPunct{\mcitedefaultmidpunct}
{\mcitedefaultendpunct}{\mcitedefaultseppunct}\relax
\EndOfBibitem
\bibitem[Gonz{\'a}lez \latin{et~al.}(2014)Gonz{\'a}lez, Landajo, and
  Presno]{gonzalez2014tracking}
Gonz{\'a}lez,~P.~F.; Landajo,~M.; Presno,~M.~J. Tracking European Union CO2
  emissions through LMDI (logarithmic-mean Divisia index) decomposition. The
  activity revaluation approach. \emph{Energy} \textbf{2014}, \emph{73},
  741--750\relax
\mciteBstWouldAddEndPuncttrue
\mciteSetBstMidEndSepPunct{\mcitedefaultmidpunct}
{\mcitedefaultendpunct}{\mcitedefaultseppunct}\relax
\EndOfBibitem
\bibitem[Wang \latin{et~al.}(2005)Wang, Chen, and Zou]{wang2005decomposition}
Wang,~C.; Chen,~J.; Zou,~J. Decomposition of energy-related CO2 emission in
  China: 1957--2000. \emph{Energy} \textbf{2005}, \emph{30}, 73--83\relax
\mciteBstWouldAddEndPuncttrue
\mciteSetBstMidEndSepPunct{\mcitedefaultmidpunct}
{\mcitedefaultendpunct}{\mcitedefaultseppunct}\relax
\EndOfBibitem
\bibitem[Wang \latin{et~al.}(2013)Wang, Zhao, Li, Liu, and
  Liang]{wang2013carbon}
Wang,~Y.; Zhao,~H.; Li,~L.; Liu,~Z.; Liang,~S. Carbon dioxide emission drivers
  for a typical metropolis using input--output structural decomposition
  analysis. \emph{Energy Policy} \textbf{2013}, \emph{58}, 312--318\relax
\mciteBstWouldAddEndPuncttrue
\mciteSetBstMidEndSepPunct{\mcitedefaultmidpunct}
{\mcitedefaultendpunct}{\mcitedefaultseppunct}\relax
\EndOfBibitem
\bibitem[Wei \latin{et~al.}(2017)Wei, Huang, Yang, Li, Hu, and
  Zhang]{wei2017driving}
Wei,~J.; Huang,~K.; Yang,~S.; Li,~Y.; Hu,~T.; Zhang,~Y. Driving forces analysis
  of energy-related carbon dioxide (CO2) emissions in Beijing: an input--output
  structural decomposition analysis. \emph{Journal of Cleaner Production}
  \textbf{2017}, \emph{163}, 58--68\relax
\mciteBstWouldAddEndPuncttrue
\mciteSetBstMidEndSepPunct{\mcitedefaultmidpunct}
{\mcitedefaultendpunct}{\mcitedefaultseppunct}\relax
\EndOfBibitem
\bibitem[Kumbaro{\u{g}}lu(2011)]{kumbarouglu2011sectoral}
Kumbaro{\u{g}}lu,~G. A sectoral decomposition analysis of Turkish CO2 emissions
  over 1990--2007. \emph{Energy} \textbf{2011}, \emph{36}, 2419--2433\relax
\mciteBstWouldAddEndPuncttrue
\mciteSetBstMidEndSepPunct{\mcitedefaultmidpunct}
{\mcitedefaultendpunct}{\mcitedefaultseppunct}\relax
\EndOfBibitem
\bibitem[Ang and Choi(1997)Ang, and Choi]{ang1997decomposition}
Ang,~B.~W.; Choi,~K.-H. Decomposition of aggregate energy and gas emission
  intensities for industry: a refined Divisia index method. \emph{The Energy
  Journal} \textbf{1997}, \emph{18}\relax
\mciteBstWouldAddEndPuncttrue
\mciteSetBstMidEndSepPunct{\mcitedefaultmidpunct}
{\mcitedefaultendpunct}{\mcitedefaultseppunct}\relax
\EndOfBibitem
\bibitem[Ang \latin{et~al.}(2003)Ang, Liu, and Chew]{ang2003perfect}
Ang,~B.~W.; Liu,~F.; Chew,~E.~P. Perfect decomposition techniques in energy and
  environmental analysis. \emph{Energy Policy} \textbf{2003}, \emph{31},
  1561--1566\relax
\mciteBstWouldAddEndPuncttrue
\mciteSetBstMidEndSepPunct{\mcitedefaultmidpunct}
{\mcitedefaultendpunct}{\mcitedefaultseppunct}\relax
\EndOfBibitem
\bibitem[Shahbaz \latin{et~al.}(2012)Shahbaz, Lean, and
  Shabbir]{shahbaz2012environmental}
Shahbaz,~M.; Lean,~H.~H.; Shabbir,~M.~S. Environmental Kuznets curve hypothesis
  in Pakistan: cointegration and Granger causality. \emph{Renewable and
  Sustainable Energy Reviews} \textbf{2012}, \emph{16}, 2947--2953\relax
\mciteBstWouldAddEndPuncttrue
\mciteSetBstMidEndSepPunct{\mcitedefaultmidpunct}
{\mcitedefaultendpunct}{\mcitedefaultseppunct}\relax
\EndOfBibitem
\bibitem[Jalil and Mahmud(2009)Jalil, and Mahmud]{jalil2009environment}
Jalil,~A.; Mahmud,~S.~F. Environment Kuznets curve for CO2 emissions: a
  cointegration analysis for China. \emph{Energy policy} \textbf{2009},
  \emph{37}, 5167--5172\relax
\mciteBstWouldAddEndPuncttrue
\mciteSetBstMidEndSepPunct{\mcitedefaultmidpunct}
{\mcitedefaultendpunct}{\mcitedefaultseppunct}\relax
\EndOfBibitem
\bibitem[Rahman and Kashem(2017)Rahman, and Kashem]{rahman2017carbon}
Rahman,~M.~M.; Kashem,~M.~A. Carbon emissions, energy consumption and
  industrial growth in Bangladesh: Empirical evidence from ARDL cointegration
  and Granger causality analysis. \emph{Energy policy} \textbf{2017},
  \emph{110}, 600--608\relax
\mciteBstWouldAddEndPuncttrue
\mciteSetBstMidEndSepPunct{\mcitedefaultmidpunct}
{\mcitedefaultendpunct}{\mcitedefaultseppunct}\relax
\EndOfBibitem
\bibitem[Pearson(1901)]{pearson1901liii}
Pearson,~K. LIII. On lines and planes of closest fit to systems of points in
  space. \emph{The London, Edinburgh, and Dublin philosophical magazine and
  journal of science} \textbf{1901}, \emph{2}, 559--572\relax
\mciteBstWouldAddEndPuncttrue
\mciteSetBstMidEndSepPunct{\mcitedefaultmidpunct}
{\mcitedefaultendpunct}{\mcitedefaultseppunct}\relax
\EndOfBibitem
\bibitem[Hotelling(1933)]{hotelling1933analysis}
Hotelling,~H. Analysis of a complex of statistical variables into principal
  components. \emph{Journal of educational psychology} \textbf{1933},
  \emph{24}, 417\relax
\mciteBstWouldAddEndPuncttrue
\mciteSetBstMidEndSepPunct{\mcitedefaultmidpunct}
{\mcitedefaultendpunct}{\mcitedefaultseppunct}\relax
\EndOfBibitem
\bibitem[Ciulla and D'Amico(2019)Ciulla, and D'Amico]{ciulla2019building}
Ciulla,~G.; D'Amico,~A. Building energy performance forecasting: A multiple
  linear regression approach. \emph{Applied Energy} \textbf{2019}, \emph{253},
  113500\relax
\mciteBstWouldAddEndPuncttrue
\mciteSetBstMidEndSepPunct{\mcitedefaultmidpunct}
{\mcitedefaultendpunct}{\mcitedefaultseppunct}\relax
\EndOfBibitem
\bibitem[Hosseini \latin{et~al.}(2019)Hosseini, Saifoddin, Shirmohammadi, and
  Aslani]{hosseini2019forecasting}
Hosseini,~S.~M.; Saifoddin,~A.; Shirmohammadi,~R.; Aslani,~A. Forecasting of
  CO2 emissions in Iran based on time series and regression analysis.
  \emph{Energy Reports} \textbf{2019}, \emph{5}, 619--631\relax
\mciteBstWouldAddEndPuncttrue
\mciteSetBstMidEndSepPunct{\mcitedefaultmidpunct}
{\mcitedefaultendpunct}{\mcitedefaultseppunct}\relax
\EndOfBibitem
\bibitem[Ostertagov{\'a}(2012)]{ostertagova2012modelling}
Ostertagov{\'a},~E. Modelling using polynomial regression. \emph{Procedia
  Engineering} \textbf{2012}, \emph{48}, 500--506\relax
\mciteBstWouldAddEndPuncttrue
\mciteSetBstMidEndSepPunct{\mcitedefaultmidpunct}
{\mcitedefaultendpunct}{\mcitedefaultseppunct}\relax
\EndOfBibitem
\bibitem[Qiao \latin{et~al.}(2020)Qiao, Lu, Zhou, Azimi, Yang, and
  Tian]{qiao2020hybrid}
Qiao,~W.; Lu,~H.; Zhou,~G.; Azimi,~M.; Yang,~Q.; Tian,~W. A hybrid algorithm
  for carbon dioxide emissions forecasting based on improved lion swarm
  optimizer. \emph{Journal of Cleaner Production} \textbf{2020}, \emph{244},
  118612\relax
\mciteBstWouldAddEndPuncttrue
\mciteSetBstMidEndSepPunct{\mcitedefaultmidpunct}
{\mcitedefaultendpunct}{\mcitedefaultseppunct}\relax
\EndOfBibitem
\bibitem[Sun \latin{et~al.}(2017)Sun, Wang, and Zhang]{sun2017factor}
Sun,~W.; Wang,~C.; Zhang,~C. Factor analysis and forecasting of CO2 emissions
  in Hebei, using extreme learning machine based on particle swarm
  optimization. \emph{Journal of cleaner production} \textbf{2017}, \emph{162},
  1095--1101\relax
\mciteBstWouldAddEndPuncttrue
\mciteSetBstMidEndSepPunct{\mcitedefaultmidpunct}
{\mcitedefaultendpunct}{\mcitedefaultseppunct}\relax
\EndOfBibitem
\bibitem[Gallo \latin{et~al.}(2014)Gallo, Conto, and Fiore]{gallo2014neural}
Gallo,~C.; Conto,~F.; Fiore,~M. A neural network model for forecasting CO2
  emission. \emph{AGRIS on-line Papers in Economics and Informatics}
  \textbf{2014}, \emph{6}, 31--36\relax
\mciteBstWouldAddEndPuncttrue
\mciteSetBstMidEndSepPunct{\mcitedefaultmidpunct}
{\mcitedefaultendpunct}{\mcitedefaultseppunct}\relax
\EndOfBibitem
\bibitem[Pino-Mej{\'\i}as \latin{et~al.}(2017)Pino-Mej{\'\i}as,
  P{\'e}rez-Fargallo, Rubio-Bellido, and Pulido-Arcas]{pino2017comparison}
Pino-Mej{\'\i}as,~R.; P{\'e}rez-Fargallo,~A.; Rubio-Bellido,~C.;
  Pulido-Arcas,~J.~A. Comparison of linear regression and artificial neural
  networks models to predict heating and cooling energy demand, energy
  consumption and CO2 emissions. \emph{Energy} \textbf{2017}, \emph{118},
  24--36\relax
\mciteBstWouldAddEndPuncttrue
\mciteSetBstMidEndSepPunct{\mcitedefaultmidpunct}
{\mcitedefaultendpunct}{\mcitedefaultseppunct}\relax
\EndOfBibitem
\bibitem[Liu \latin{et~al.}(2018)Liu, Bao, Wang, and Zhang]{liu2018hybrid}
Liu,~T.; Bao,~J.; Wang,~J.; Zhang,~Y. A hybrid CNN--LSTM algorithm for online
  defect recognition of CO2 welding. \emph{Sensors} \textbf{2018}, \emph{18},
  4369\relax
\mciteBstWouldAddEndPuncttrue
\mciteSetBstMidEndSepPunct{\mcitedefaultmidpunct}
{\mcitedefaultendpunct}{\mcitedefaultseppunct}\relax
\EndOfBibitem
\bibitem[Zuo \latin{et~al.}(2020)Zuo, Guo, and Cheng]{zuo2020lstm}
Zuo,~Z.; Guo,~H.; Cheng,~J. An LSTM-STRIPAT model analysis of China’s 2030
  CO2 emissions peak. \emph{Carbon Management} \textbf{2020}, \emph{11},
  577--592\relax
\mciteBstWouldAddEndPuncttrue
\mciteSetBstMidEndSepPunct{\mcitedefaultmidpunct}
{\mcitedefaultendpunct}{\mcitedefaultseppunct}\relax
\EndOfBibitem
\bibitem[Zhu \latin{et~al.}(2022)Zhu, Al-Ahmed, Shakir, and
  Olszewska]{zhu2022lstm}
Zhu,~Y.; Al-Ahmed,~S.~A.; Shakir,~M.~Z.; Olszewska,~J.~I. LSTM-based
  IoT-enabled CO2 steady-state forecasting for indoor air quality monitoring.
  \emph{Electronics} \textbf{2022}, \emph{12}, 107\relax
\mciteBstWouldAddEndPuncttrue
\mciteSetBstMidEndSepPunct{\mcitedefaultmidpunct}
{\mcitedefaultendpunct}{\mcitedefaultseppunct}\relax
\EndOfBibitem
\bibitem[Kumari and Singh(2022)Kumari, and Singh]{kumari2022machine}
Kumari,~S.; Singh,~S.~K. Machine learning-based time series models for
  effective CO2 emission prediction in India. \emph{Environmental Science and
  Pollution Research} \textbf{2022}, 1--16\relax
\mciteBstWouldAddEndPuncttrue
\mciteSetBstMidEndSepPunct{\mcitedefaultmidpunct}
{\mcitedefaultendpunct}{\mcitedefaultseppunct}\relax
\EndOfBibitem
\bibitem[Liu \latin{et~al.}(2020)Liu, Ciais, Deng, Davis, Zheng, Wang, Cui,
  Zhu, Dou, Ke, \latin{et~al.} others]{liu2020carbon}
Liu,~Z.; Ciais,~P.; Deng,~Z.; Davis,~S.~J.; Zheng,~B.; Wang,~Y.; Cui,~D.;
  Zhu,~B.; Dou,~X.; Ke,~P., \latin{et~al.}  Carbon Monitor, a near-real-time
  daily dataset of global CO2 emission from fossil fuel and cement production.
  \emph{Scientific data} \textbf{2020}, \emph{7}, 392\relax
\mciteBstWouldAddEndPuncttrue
\mciteSetBstMidEndSepPunct{\mcitedefaultmidpunct}
{\mcitedefaultendpunct}{\mcitedefaultseppunct}\relax
\EndOfBibitem
\bibitem[Ke \latin{et~al.}(2023)Ke, Deng, Zhu, Zheng, Wang, Boucher, Arous,
  Zhou, Andrew, Dou, \latin{et~al.} others]{ke2023carbon}
Ke,~P.; Deng,~Z.; Zhu,~B.; Zheng,~B.; Wang,~Y.; Boucher,~O.; Arous,~S.~B.;
  Zhou,~C.; Andrew,~R.~M.; Dou,~X., \latin{et~al.}  Carbon Monitor Europe
  near-real-time daily CO2 emissions for 27 EU countries and the United
  Kingdom. \emph{Scientific Data} \textbf{2023}, \emph{10}, 374\relax
\mciteBstWouldAddEndPuncttrue
\mciteSetBstMidEndSepPunct{\mcitedefaultmidpunct}
{\mcitedefaultendpunct}{\mcitedefaultseppunct}\relax
\EndOfBibitem
\bibitem[Faruque \latin{et~al.}(2022)Faruque, Rabby, Hossain, Islam, Rashid,
  and Muyeen]{faruque2022comparative}
Faruque,~M.~O.; Rabby,~M. A.~J.; Hossain,~M.~A.; Islam,~M.~R.; Rashid,~M.
  M.~U.; Muyeen,~S. A comparative analysis to forecast carbon dioxide
  emissions. \emph{Energy Reports} \textbf{2022}, \emph{8}, 8046--8060\relax
\mciteBstWouldAddEndPuncttrue
\mciteSetBstMidEndSepPunct{\mcitedefaultmidpunct}
{\mcitedefaultendpunct}{\mcitedefaultseppunct}\relax
\EndOfBibitem
\bibitem[Zheng \latin{et~al.}(2019)Zheng, Chen, and Luo]{zheng2019spatial}
Zheng,~Z.; Chen,~H.; Luo,~X. Spatial granularity analysis on electricity
  consumption prediction using LSTM recurrent neural network. \emph{Energy
  Procedia} \textbf{2019}, \emph{158}, 2713--2718\relax
\mciteBstWouldAddEndPuncttrue
\mciteSetBstMidEndSepPunct{\mcitedefaultmidpunct}
{\mcitedefaultendpunct}{\mcitedefaultseppunct}\relax
\EndOfBibitem
\bibitem[Mir \latin{et~al.}(2022)Mir, Yadav, and Singh]{VKY0}
Mir,~S.~H.; Yadav,~V.~K.; Singh,~J.~K. Efficient CO2 capture and activation on
  novel two-dimensional transition metal borides. \emph{ACS Applied Materials
  \& Interfaces} \textbf{2022}, \emph{14}, 29703--29710\relax
\mciteBstWouldAddEndPuncttrue
\mciteSetBstMidEndSepPunct{\mcitedefaultmidpunct}
{\mcitedefaultendpunct}{\mcitedefaultseppunct}\relax
\EndOfBibitem
\bibitem[Wang \latin{et~al.}(2019)Wang, Ye, Gong, Wu, Miao, Tada, and
  Hosono]{VKY1}
Wang,~J.; Ye,~T.-N.; Gong,~Y.; Wu,~J.; Miao,~N.; Tada,~T.; Hosono,~H. Discovery
  of hexagonal ternary phase Ti2InB2 and its evolution to layered boride TiB.
  \emph{Nature communications} \textbf{2019}, \emph{10}, 2284\relax
\mciteBstWouldAddEndPuncttrue
\mciteSetBstMidEndSepPunct{\mcitedefaultmidpunct}
{\mcitedefaultendpunct}{\mcitedefaultseppunct}\relax
\EndOfBibitem
\bibitem[Ma \latin{et~al.}(2022)Ma, Wang, Li, Li, and Fan]{VKY2}
Ma,~N.; Wang,~T.; Li,~N.; Li,~Y.; Fan,~J. New phases of MBenes M2B (M= Sc, Ti,
  and V) as high-capacity electrode materials for rechargeable magnesium ion
  batteries. \emph{Applied Surface Science} \textbf{2022}, \emph{571},
  151275\relax
\mciteBstWouldAddEndPuncttrue
\mciteSetBstMidEndSepPunct{\mcitedefaultmidpunct}
{\mcitedefaultendpunct}{\mcitedefaultseppunct}\relax
\EndOfBibitem
\bibitem[Li \latin{et~al.}(2018)Li, Zhang, Guo, Huan, Xi, Li, Bai, and
  Yan]{VKY3}
Li,~M.; Zhang,~R.; Guo,~Y.; Huan,~Y.; Xi,~J.; Li,~Y.; Bai,~Z.; Yan,~X.
  Introducing lead acetate into stoichiometric perovskite lewis acid-base
  precursor for improved solar cell photovoltaic performance. \emph{Journal of
  Alloys and Compounds} \textbf{2018}, \emph{767}, 829--837\relax
\mciteBstWouldAddEndPuncttrue
\mciteSetBstMidEndSepPunct{\mcitedefaultmidpunct}
{\mcitedefaultendpunct}{\mcitedefaultseppunct}\relax
\EndOfBibitem
\bibitem[Zhang \latin{et~al.}(2019)Zhang, Dai, Xiang, Wang, Zhang, and
  Zhou]{VKY4}
Zhang,~H.; Dai,~F.-Z.; Xiang,~H.; Wang,~X.; Zhang,~Z.; Zhou,~Y. Phase pure and
  well crystalline Cr2AlB2: A key precursor for two-dimensional CrB.
  \emph{Journal of Materials Science \& Technology} \textbf{2019}, \emph{35},
  1593--1600\relax
\mciteBstWouldAddEndPuncttrue
\mciteSetBstMidEndSepPunct{\mcitedefaultmidpunct}
{\mcitedefaultendpunct}{\mcitedefaultseppunct}\relax
\EndOfBibitem
\bibitem[Wang \latin{et~al.}(2022)Wang, Guo, Zhang, Li, Zhao, and Wang]{VKY5}
Wang,~Z.; Guo,~Y.; Zhang,~Q.; Li,~Z.; Zhao,~Y.; Wang,~H. Alkanolamine
  intercalation assisted liquid phase exfoliation of titanium carbide MXene
  nanosheets for highly efficient photocatalytic CO2 reduction. \emph{Journal
  of Molecular Liquids} \textbf{2022}, \emph{367}, 120578\relax
\mciteBstWouldAddEndPuncttrue
\mciteSetBstMidEndSepPunct{\mcitedefaultmidpunct}
{\mcitedefaultendpunct}{\mcitedefaultseppunct}\relax
\EndOfBibitem
\bibitem[Mou \latin{et~al.}(2023)Mou, Li, Zhang, Xu, Fan, and Bei]{VKY6}
Mou,~J.; Li,~S.; Zhang,~W.; Xu,~W.; Fan,~S.; Bei,~G. Deintercalation of Al from
  MoAlB by molten salt etching to achieve a Mo 2 AlB 2 compound and 2D MoB
  nanosheets. \emph{Journal of Advanced Ceramics} \textbf{2023}, \emph{12},
  943--953\relax
\mciteBstWouldAddEndPuncttrue
\mciteSetBstMidEndSepPunct{\mcitedefaultmidpunct}
{\mcitedefaultendpunct}{\mcitedefaultseppunct}\relax
\EndOfBibitem
\bibitem[Weerasinghe \latin{et~al.}(2023)Weerasinghe, Wu, Lee, Lin, Anariba,
  Li, Seng, Sim, and Wu]{VKY7}
Weerasinghe,~P. V.~T.; Wu,~S.; Lee,~W.~C.; Lin,~M.; Anariba,~F.; Li,~X.;
  Seng,~D. H.~L.; Sim,~J.~Y.; Wu,~P. Efficient Synthesis of 2D Mica Nanosheets
  by Solvothermal and Microwave-Assisted Techniques for CO2 Capture
  Applications. \emph{Materials} \textbf{2023}, \emph{16}, 2921\relax
\mciteBstWouldAddEndPuncttrue
\mciteSetBstMidEndSepPunct{\mcitedefaultmidpunct}
{\mcitedefaultendpunct}{\mcitedefaultseppunct}\relax
\EndOfBibitem
\bibitem[Xiao and Shen(2021)Xiao, and Shen]{VKY8}
Xiao,~Y.; Shen,~C. Transition-metal borides (MBenes) as new high-efficiency
  catalysts for nitric oxide electroreduction to ammonia by a high-throughput
  approach. \emph{Small} \textbf{2021}, \emph{17}, 2100776\relax
\mciteBstWouldAddEndPuncttrue
\mciteSetBstMidEndSepPunct{\mcitedefaultmidpunct}
{\mcitedefaultendpunct}{\mcitedefaultseppunct}\relax
\EndOfBibitem
\bibitem[Hochreiter and Schmidhuber(1997)Hochreiter, and
  Schmidhuber]{hochreiter1997long}
Hochreiter,~S.; Schmidhuber,~J. Long short-term memory. \emph{Neural
  computation} \textbf{1997}, \emph{9}, 1735--1780\relax
\mciteBstWouldAddEndPuncttrue
\mciteSetBstMidEndSepPunct{\mcitedefaultmidpunct}
{\mcitedefaultendpunct}{\mcitedefaultseppunct}\relax
\EndOfBibitem
\bibitem[Van~Houdt \latin{et~al.}(2020)Van~Houdt, Mosquera, and
  N{\'a}poles]{van2020review}
Van~Houdt,~G.; Mosquera,~C.; N{\'a}poles,~G. A review on the long short-term
  memory model. \emph{Artificial Intelligence Review} \textbf{2020}, \emph{53},
  5929--5955\relax
\mciteBstWouldAddEndPuncttrue
\mciteSetBstMidEndSepPunct{\mcitedefaultmidpunct}
{\mcitedefaultendpunct}{\mcitedefaultseppunct}\relax
\EndOfBibitem
\bibitem[Graves and Graves(2012)Graves, and Graves]{graves2012long}
Graves,~A.; Graves,~A. Long short-term memory. \emph{Supervised sequence
  labelling with recurrent neural networks} \textbf{2012}, 37--45\relax
\mciteBstWouldAddEndPuncttrue
\mciteSetBstMidEndSepPunct{\mcitedefaultmidpunct}
{\mcitedefaultendpunct}{\mcitedefaultseppunct}\relax
\EndOfBibitem
\bibitem[A{\u{g}}bulut \latin{et~al.}(2021)A{\u{g}}bulut, G{\"u}rel, and
  Bi{\c{c}}en]{augbulut2021prediction}
A{\u{g}}bulut,~{\"U}.; G{\"u}rel,~A.~E.; Bi{\c{c}}en,~Y. Prediction of daily
  global solar radiation using different machine learning algorithms:
  Evaluation and comparison. \emph{Renewable and Sustainable Energy Reviews}
  \textbf{2021}, \emph{135}, 110114\relax
\mciteBstWouldAddEndPuncttrue
\mciteSetBstMidEndSepPunct{\mcitedefaultmidpunct}
{\mcitedefaultendpunct}{\mcitedefaultseppunct}\relax
\EndOfBibitem
\bibitem[Kumar \latin{et~al.}(2020)Kumar, Kumar, and Kumar]{kumar2020time}
Kumar,~R.; Kumar,~P.; Kumar,~Y. Time series data prediction using IoT and
  machine learning technique. \emph{Procedia computer science} \textbf{2020},
  \emph{167}, 373--381\relax
\mciteBstWouldAddEndPuncttrue
\mciteSetBstMidEndSepPunct{\mcitedefaultmidpunct}
{\mcitedefaultendpunct}{\mcitedefaultseppunct}\relax
\EndOfBibitem
\bibitem[A{\u{g}}bulut \latin{et~al.}(2021)A{\u{g}}bulut, G{\"u}rel, and
  Sar{\i}demir]{augbulut2021experimental}
A{\u{g}}bulut,~{\"U}.; G{\"u}rel,~A.~E.; Sar{\i}demir,~S. Experimental
  investigation and prediction of performance and emission responses of a CI
  engine fuelled with different metal-oxide based nanoparticles--diesel blends
  using different machine learning algorithms. \emph{Energy} \textbf{2021},
  \emph{215}, 119076\relax
\mciteBstWouldAddEndPuncttrue
\mciteSetBstMidEndSepPunct{\mcitedefaultmidpunct}
{\mcitedefaultendpunct}{\mcitedefaultseppunct}\relax
\EndOfBibitem
\bibitem[Giannozzi \latin{et~al.}(2009)Giannozzi, Baroni, Bonini, Calandra,
  Car, Cavazzoni, Ceresoli, Chiarotti, Cococcioni, Dabo, \latin{et~al.}
  others]{VKY9}
Giannozzi,~P.; Baroni,~S.; Bonini,~N.; Calandra,~M.; Car,~R.; Cavazzoni,~C.;
  Ceresoli,~D.; Chiarotti,~G.~L.; Cococcioni,~M.; Dabo,~I., \latin{et~al.}
  QUANTUM ESPRESSO: a modular and open-source software project for quantum
  simulations of materials. \emph{Journal of physics: Condensed matter}
  \textbf{2009}, \emph{21}, 395502\relax
\mciteBstWouldAddEndPuncttrue
\mciteSetBstMidEndSepPunct{\mcitedefaultmidpunct}
{\mcitedefaultendpunct}{\mcitedefaultseppunct}\relax
\EndOfBibitem
\bibitem[Perdew \latin{et~al.}(1996)Perdew, Burke, Ernzerhof, \latin{et~al.}
  others]{VKY10}
Perdew,~J.; Burke,~K.; Ernzerhof,~M., \latin{et~al.}  NOL 70118 J. Quantum
  theory group tulane university. \emph{Phys. Rev. Lett.} \textbf{1996},
  \emph{77}, 3865--3868\relax
\mciteBstWouldAddEndPuncttrue
\mciteSetBstMidEndSepPunct{\mcitedefaultmidpunct}
{\mcitedefaultendpunct}{\mcitedefaultseppunct}\relax
\EndOfBibitem
\bibitem[Ehrlich \latin{et~al.}(2011)Ehrlich, Moellmann, Reckien, Bredow, and
  Grimme]{VKY11}
Ehrlich,~S.; Moellmann,~J.; Reckien,~W.; Bredow,~T.; Grimme,~S.
  System-dependent dispersion coefficients for the DFT-D3 treatment of
  adsorption processes on ionic surfaces. \emph{ChemPhysChem} \textbf{2011},
  \emph{12}, 3414--3420\relax
\mciteBstWouldAddEndPuncttrue
\mciteSetBstMidEndSepPunct{\mcitedefaultmidpunct}
{\mcitedefaultendpunct}{\mcitedefaultseppunct}\relax
\EndOfBibitem
\bibitem[Monkhorst and Pack(1976)Monkhorst, and Pack]{VKY12}
Monkhorst,~H.~J.; Pack,~J.~D. Special points for Brillouin-zone integrations.
  \emph{Physical review B} \textbf{1976}, \emph{13}, 5188\relax
\mciteBstWouldAddEndPuncttrue
\mciteSetBstMidEndSepPunct{\mcitedefaultmidpunct}
{\mcitedefaultendpunct}{\mcitedefaultseppunct}\relax
\EndOfBibitem
\bibitem[Cousineau and Chartier(2010)Cousineau, and
  Chartier]{cousineau2010outliers}
Cousineau,~D.; Chartier,~S. Outliers detection and treatment: a review.
  \emph{International Journal of Psychological Research} \textbf{2010},
  \emph{3}, 58--67\relax
\mciteBstWouldAddEndPuncttrue
\mciteSetBstMidEndSepPunct{\mcitedefaultmidpunct}
{\mcitedefaultendpunct}{\mcitedefaultseppunct}\relax
\EndOfBibitem
\bibitem[Tripathy(2013)]{tripathy2013comparison}
Tripathy,~S.~S. Comparison of statistical methods for outlier detection in
  proficiency testing data on analysis of lead in aqueous solution.
  \emph{American Journal of Theoretical and Applied Statistics} \textbf{2013},
  \emph{2}, 233\relax
\mciteBstWouldAddEndPuncttrue
\mciteSetBstMidEndSepPunct{\mcitedefaultmidpunct}
{\mcitedefaultendpunct}{\mcitedefaultseppunct}\relax
\EndOfBibitem
\bibitem[Sandbhor and Chaphalkar(2019)Sandbhor, and
  Chaphalkar]{sandbhor2019impact}
Sandbhor,~S.; Chaphalkar,~N. Impact of outlier detection on neural networks
  based property value prediction. Information Systems Design and Intelligent
  Applications: Proceedings of Fifth International Conference INDIA 2018 Volume
  1. 2019; pp 481--495\relax
\mciteBstWouldAddEndPuncttrue
\mciteSetBstMidEndSepPunct{\mcitedefaultmidpunct}
{\mcitedefaultendpunct}{\mcitedefaultseppunct}\relax
\EndOfBibitem
\end{mcitethebibliography}
\end{document}